 \documentclass[final,5p,times,twocolumn]{elsarticle}
\usepackage{graphicx}
\usepackage{stfloats}
\usepackage{amssymb}
\usepackage{amsmath}
\usepackage{color}
\usepackage{algorithm}
\usepackage{algorithmic}
\usepackage{graphicx}
\usepackage{xspace}

\usepackage{natbib}

\usepackage{lineno,hyperref}

\DeclareMathAlphabet{\mathpzc}{OT1}{pzc}{m}{it}

\newcommand{\mathkomma}{\quad ,}
\newcommand{\mathpunkt}{\quad .}

\def\eg{e.g.\@\xspace}
\def\ie{i.e.\@\xspace}

\newcommand{\red}[1]{\textcolor{black}{#1}}

\usepackage{multirow} 
\usepackage{booktabs}
\usepackage{array,graphicx} 
\usepackage{pifont}
\newcommand*\OK{\ding{51}}

\newcommand{\mcrot}[4]{\multicolumn{#1}{#2}{\rlap{\rotatebox{#3}{#4}~}}} 

\newcommand*{\twoelementtable}[3][l]%
{%
    \begin{tabular}[t]{@{}#1@{}}%
        #2\tabularnewline
        #3%
    \end{tabular}%
}

\journal{IJCV}
\begin{document}  

\begin{frontmatter}

\title{Semantic Decomposition and Recognition \\of  Long and Complex Manipulation Action Sequences}

\author{Eren Erdal Aksoy$^{\star,1}$} 
\author{Adil Orhan$^{1}$}
\author{Florentin W\"org\"otter$^{2}$}
\address{$^{1}$Institute for Anthropomatics and Robotics, \\
              High Performance Humanoid Technologies (H$^{2}$T), \\
              Karlsruhe Institute of Technology, Germany }
\address{$^{2}$Georg-August-Universit\"at G\"ottingen, BCCN \\ Inst. Physics-3, Friedrich-Hund Platz 1, D-37077 G{\"o}ttingen, Germany}




\begin{abstract}

Understanding   continuous human actions is a non-trivial but important problem in computer vision.
Although there exists a large corpus of work in the recognition of action sequences,  most approaches suffer from problems relating to vast variations in motions, action combinations, and scene contexts.
In this paper, we introduce a novel method for semantic segmentation and recognition of long and complex manipulation action tasks, such as ``{\it preparing a breakfast}" or ``{\it making a sandwich}". We represent  manipulations with our recently introduced ``Semantic Event Chain" (SEC) concept, which captures the underlying spatiotemporal structure of an action invariant to motion, velocity, and scene context.
Solely based on the spatiotemporal interactions between manipulated objects and hands in the extracted SEC, the framework automatically parses individual manipulation streams performed either sequentially or concurrently.
Using event chains, our method further extracts basic primitive elements of each parsed manipulation. Without requiring any prior object knowledge, the proposed framework can also extract object-like scene entities that exhibit the same role in semantically similar manipulations.
We conduct extensive experiments on various recent  datasets to validate the robustness of the framework.

\end{abstract}


\end{frontmatter}

\section{Introduction}

Humans are  able to understand a very large variety of complex actions performed by others. Automatic monitoring of human actions   is, on the other hand, a long-standing and challenging problem in computer vision. In the literature, one finds substantial efforts along the lines of temporal segmentation and recognition of continuous human action sequences \citep{Yamato1992,Rui00,Gupta2009,Wang2012,Zhou13,Koppula13}.
Recent works have mostly approached this problem from the perspective of analyzing  motion patterns and by matching  appearance-based features for the monitoring  of action sequences. Due to the very large intra-person motion variability, such approaches, however, require fully labeled large training data and do not generalize well.

Different from conventional approaches, we here introduce a novel method for action understanding, that relies only on the spatiotemporal hand-object relations  that happen during an action. We use our recently introduced  ``Semantic Event Chain" (SEC) concept  \citep{Aksoy2010,Aksoy2011} as a descriptive action representation method.
SECs capture the underlying spatiotemporal structure of continuous actions by sampling only decisive key temporal points derived from the spatial interactions between hands and objects in the scene. The SEC representation is invariant to large variations in   trajectory, velocity, object type and pose used in the action. Therefore, SECs can be employed for the  classification task of actions as  demonstrated in various experiments in \citep{Aksoy2010,Aksoy2011,Aksoy2013}.
In this paper, we  aim at analyzing long and complex action sequences in which a human is manipulating multiple objects in different orders for a specific task, such as ``{\it preparing a breakfast}" or ``{\it making a sandwich}". Such actions are commonly called ``{\it manipulation}" since  hands are intensively interacting with objects towards a goal. Thus, instead of analysing entire human body configurations or motions, we  only (compactly) encode spatiotemporal hand-object relations by those event chains.

\begin{figure}[!t]
\centering
 \includegraphics[scale=0.8]{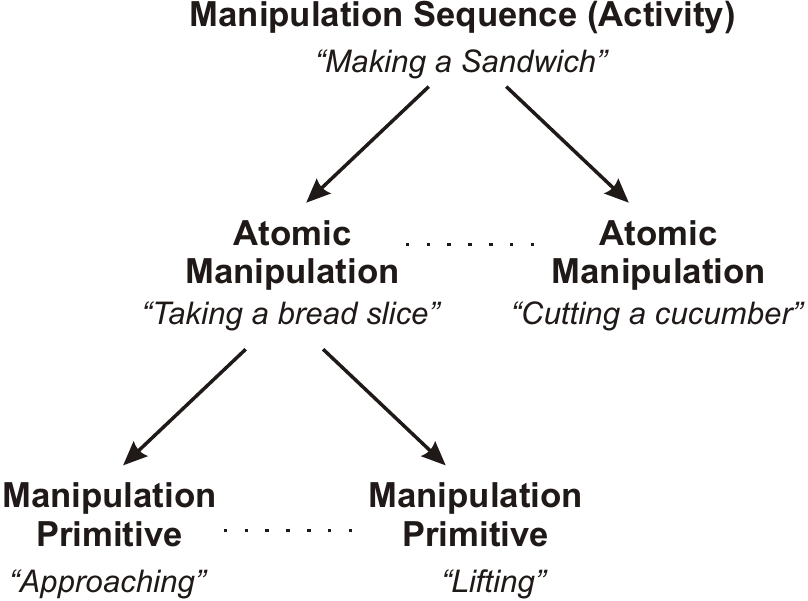}
  \caption{Taxonomy of manipulation actions. }
\label{fig:action_taxonomy}
\end{figure}

In the context of action understanding, different taxonomies have been proposed to date in the literature. The term {\it action} is a rather general description for any type of individual behavior like {\it walking}, {\it jumping}, or {\it pushing}. In the context of this paper, more specific terms such as {\it manipulation} or {\it manipulation action} are used denoting that we are dealing which specific actions where hands are interacting with objects.  Fig.~\ref{fig:action_taxonomy} shows the hierarchy of terms used in this paper. A manipulation primitive, \eg {\it approach} or {\it lift}, is the smallest basic component of a manipulation. Different sequences of primitives lead to different types of atomic manipulations  such as {\it pushing} or {\it cutting}. Finally, manipulation sequences or activities, \eg ``{\it making a sandwich}", contain a series of chained atomic manipulations. We note that a semantic understanding of actions can happen for any of these components but sometimes also ``beneath", for example at the level of ``motions" (\eg moving your hand in a certain way to perform ``punch" or ``push"). Both would amount to an ``Approach" (bottom of Fig.~\ref{fig:action_taxonomy}) but such (dynamic) levels are not included in this taxonomy and the SEC framework represents one certain specific level by which many manipulation actions can be distinguished but certain other ones will be considered type-identical. We refer the reader to the Discussion section, where we will dig a bit deeper into these aspects, which -- albeit being even of a philosophical origin -- have quite a strong influence on the algorithmic treatment of the ``action problem".

The proposed framework has two processing stages: manipulation {\it temporal segmentation} and {\it recognition}. The temporal segmentation stage detects the changes in the spatiotemporal relations emerge between objects and hands in the scene. These detected changes results in parsing of individual actions performed sequentially or concurrently. 
The recognition stage requires an alphabet of  atomic manipulations (\eg~ {\it cutting} or {\it stirring}) which is provided up-front by learned SEC models for each ``atom" using the unsupervised learning method introduced in \cite{AksoyRAS2015}. Using this, manipulation sequences can be recognized and the framework then also deals with objects.

However, roles of individual objects are usually not unique.
A cup, for example, is commonly used for {\it filling}  and/or {\it drinking} actions. The same cup can, however, also be utilized as a {\it pedestal} to put something on top of it after having first turned it upside down. Thus, depending on the intended goal, roles of objects can vary from manipulation  to manipulation. We will show that we can extract object-like entities (image segments) with the here proposed framework and cluster them according to their exhibited roles in each recognized manipulation type without requiring any prior object knowledge.

The rest of the paper is organized as follows. We start with introducing the state of the art. We then continue with a detailed description of each processing step. Next, we provide experimental results on various datasets and finally we finish with a discussion.

\section{State of the Art}
\label{sec:soa}

Understanding continuous human actions is of fundamental importance in computer vision and has very broad potential application areas such as video surveillance, multimedia retrieval, virtual reality, and human-robot interaction \citep{Sukthankar2007,Pardowitz08,HoaiLD2011,Pei13}. There is a large corpus of work in both temporal segmentation and recognition of human actions in computer vision and machine learning. Achievements obtained in these topics will now be  summarized but we cannot provide complete coverage of all works in these fields. See \cite{Poppe2010}, \cite{Ahad2011}, or \cite{Weinland2011} for comprehensive surveys.

\subsection{Action Temporal Segmentation}
\label{sec:actiondecomposition}

Temporal segmentation, \ie decomposition, is the process of segmenting the input data stream, \ie action sequences, into individual action instances, \ie atomic actions.

The main difficulties here are the possibly large number of action combinations, the variable durations of the different atoms, and the irregularity and variability of actions performed by different people in different (scene) contexts.
To cope with these problems, there exist different approaches such as boundary detection and sliding window methods, as well as higher-level grammars, which are widely used.

Boundary detection methods \citep{Rui00,Weinland2006,Shiv2008} essentially investigate start and end points of actions from temporal discontinuities or extrema in acceleration or velocities of the motion profiles. Although such approaches are attractive due to being invariant to action classes, they highly depend on the observed motion pattern which can exhibit high intra-class variations. Conventional sliding window approaches \citep{Zhong2004,Sukthankar2007} are searching for correspondences between previously learned action features and the current action segment under the sliding window.
The temporal segmentation performance, however, heavily depends on the recognition results which can be affected by the predefined window size.

Alternatively, higher-level grammars \citep{Peursum2004,Lv2006} build a single large network from individually modelled actions. Such grammars are then used to model transitions between single actions to further parse action sequences by computing the minimum cost path through the network using efficient dynamic programming techniques. However, such methods require large amount of training data to learn a state sequence for each action and also to capture state transitions between individual actions.

Along these lines, not only grammars with generative models, \eg Hidden Markov Models (HMMs) \citep{Peursum2004,Lv2006}, but also discriminative frameworks based on multi-class Support Vector Machines (SVM) \citep{HoaiLD2011} and semi-Markov models \citep{Qinfeng11} were meanwhile proposed  to perform simultaneous action segmentation and recognition.

Recent work \cite{Wang2012} also introduced a probabilistic graphical model with additional substructure transition and discriminative boundary models in order to tackle the problem of continuous action segmentation and recognition.
An unsupervised hierarchical bottom-up framework was presented in \cite{Zhou13} for temporal partitioning of human motion into disjoint segments.

Although all those approaches yield encouraging results, the requirement of fully labeled training data limits transfer to new sequences. Such approaches are based on bottom-up continuous motion patterns that have high variability in appearance and shape across individual demonstrations of the same action. The computational complexity, as seen in \cite{Zhou13}, also limits their applicability to long sequences.

In contrast to the aforementioned temporal segmentation approaches, we propose a method that corresponds to top-down semantic analysis of the video data without being affected by the low level data variations in object or motion domains.

Among the existing methods, the work in \cite{Pei13}, which is an event parsing approach based on a stochastic event grammar, is most strongly related to our framework since it also employs   binary spatial relations (\eg touch, near, in, etc.) between objects and agents in the scene. In contrast to our method, this framework heavily relies on a semi-supervised object recognition in order to derive atomic actions. Each action is coupled with an object, thus, a new set of atomic actions has to be learned in the case of a scenario with novel objects. Different from this, our approach does not require any prior object information and just relies on the semantic interaction between objects and hands in the scene.

\subsection{Action Recognition}
\label{sec:actionrecognition}

Action recognition is the labeling process of a given image sequence, which can be considered as a four dimensional data stream composed of spatial and temporal components.
There exists  extensive literature on topics related to action recognition. The previous works can be categorized under two main groups based on the action types. The first group of work \citep{Bobick2001,Sminchisescu2006,Scovanner2007,Laptev2008}
benefits from the intrinsic hand or body movement features, and concentrates on monitoring of full body motions, such as {\it walking} and {\it running}. The second group covers manipulation actions (\eg~{\it cutting, stirring}) in which interactions between objects and hands play the most crucial role rendering the discriminative cues. Our proposed recognition approach falls into this group, which are in general less understood and less investigated. Only a few solutions have been proposed so far \citep{Gupta2009,Fathi2011,Kjellstrom11,Yang13,Ramirez2013}.

\subsubsection{Recognition of Human Motion}
\label{sec:recognitionofhumanmotion}

Vision-based human action recognition methods have two processing stages: action representation and classification. In the action representation phase, most proposed techniques extract global or local discriminative image features either in a top-down fashion by tracking regions of interest (\eg a detected person in the scene) or as a collection of independent patches in a bottom-up fashion. Global feature representation methods encode each region of interest as a whole from silhouettes \citep{Bobick2001}, contours \citep{Chen2006}, or optical flow \citep{Efros03}. When using a local representation, however, patches are calculated around space-time interest points detected by corner detectors \citep{Laptev2003}, local histograms of oriented gradients (HOG) together with histograms of oriented flow (HOF) descriptors \citep{Laptev2008}, or scale-invariant feature transform (SIFT) descriptors \citep{Scovanner2007}. In the action classification phase, approaches  are mostly based on generative or discriminative temporal state-space models. Generative approaches (\eg HMMs \citep{Yamato1992,Hoang2012}) learn to model  each action class with all variations, whereas discriminative models (\eg Conditional Random Fields (CRF) \citep{Sminchisescu2006,Kjellstrom11}) learn the probability of each action conditioned on the observations without modeling the class.

Although such probabilistic frameworks have by far been most widely used for action recognition, they heavily rely on low-level scene features together with the body or hand motion  without employing any semantic information.

In addition, classical HMM based approaches are not suitable for recognizing parallel streams of actions  \citep{Graf2010} and cannot easily describe structures  with repetitions or recursions \citep{Lee2013}. In contrast to generative HMM based frameworks, our event chain based action representation method also obeys the Markovian assumption, but the main difference is that all states, \ie columns, in the event chains are observable. These states represent {\it key events}, \ie topological changes in the scene. Furthermore, since detailed movement variations are not considered, event chains do not require a large corpus of training data for learning individual actions as shown in our previous works \citep{Aksoy2011,AksoyRAS2015}.

\subsubsection{Recognition of Manipulation Actions}
\label{sec:recognitionofmaniac}

Ideas to utilize topological relations to reach semantics of manipulation actions can be found as early as in 1975.
The first approach, introduced in \cite{Badler1975}, represented a scene by directed graphs in which each graph node identifies one object and edges describe relative spatial information (\eg left, front, etc.) between objects. Based on object movement patterns, \ie topological changes in the scene, events are defined to represent actions. The main drawback of this approach is that actions could not really be observed by vision at this time and observation is substituted by idealized hand-made image sequences. 

In the late nineties, causal semantics has been started to be used for interpreting manipulation videos. \cite{Brand1996} analyzed globally consistent causal evolution of the scene over time. The method  detected meaningful changes in the motions and collisions of surfaces of foreground-segmented scene blobs. The same method was extended with heuristic and probabilistic models  in \cite{Brand97} to enforce longer-term consistencies in the video parse. 
\red{Different from our approach, this method requires prior object information for scene blob detection and employs motion features such as object velocity profile, which very much harms the generalization property of action recognition.
}
Along these lines, \cite{Siskind1994} presented a logical semantic notion for describing event primitives of simulated simple motions in animated line drawings. Furthermore, a maximum-likelihood-based approach was introduced in \cite{Siskind1996} to reason about a stream of 2D  ellipses, each abstractly represented the position, orientation, shape, and size of the manipulated objects in manipulation actions. 
\red{\cite{Bobick1998}  applied a stochastic Context-Free Grammar (CFG) on top of an HMM based gesture detector. The discretized HMM output was fed to the CFG parser to estimate discrete symbol streams.
}
More recently, \cite{Fern02} suggested to also incorporate force dynamic with temporal and relational information to recognize visual events in manipulation videos.
\red{\cite{Minnen2003} introduced parameterized stochastic grammars to recognize and make predictions about actions without requiring the object identity, but they were limited to recognizing semantically complex action including concurrent events. 
The work of \cite{Ryoo2006} was a hierarchical approach which started with the extraction of human body-segments and continued with the estimation of body poses at each frame. Gesture sequences were then estimated as symbolic scene states. At the highest level, a CFG was introduced to represent recursive action concepts. Although this work has similarities to our framework in terms of spatiotemporal scene representation, their framework rather focuses on human motions (\eg hugging and punching) and can not handle missing or noisy sub-events that can occur during the action. 
}

Even today there are still only a few approaches  \citep{Sridhar08,Kjellstrom11,Yang13,Nagahama2013,Ramirez2013} attempting to arrive at the semantics of manipulation actions in conjunction with assessing the manipulated objects.
\cite{Sridhar08} advocates a method for encoding an entire manipulation sequence by an activity graph, which stores the complete stream of spatiotemporal object interactions. The main difficulty here is that very complex and large activity graphs need to be decomposed for the further recognition process. In the work of \cite{Kjellstrom11}, segmented hand poses and velocities are used to classify manipulation actions. A histogram of gradients approach with a support vector machine classifier is used to categorize manipulated objects. Factorial conditional random fields are then employed to compute the correlation between objects and manipulations. However, this work does not consider interactions between the objects. Different from this, visual semantic graphs, inspired from our scene graphs, were introduced in  \cite{Yang13}  to recognize abstract action consequences (\eg~{\it Assemble}, {\it Transfer}) only based on changes in the  structure of the manipulated object without considering interactions between hands and objects. \cite{Nagahama2013} presented a method for hierarchical estimation of contact relationships (\eg~{\it on}, {\it into}) between multiple objects. The previous work in \cite{Ramirez2013} suggested extraction of abstract hand movements, such as {\it moving}, {\it not \ moving} or {\it tool used}, to further reason about more specific action primitives  (\eg~{\it Reaching}, {\it Holding})  by employing not only hand movements but also the object information. Their methods are rather for detecting actions which span only across short time intervals.
Although all those works to a certain extent improve the recognition of manipulations and/or objects, none of them addresses the temporal segmentation of long chained manipulation sequences into single atomic elements or even into key events, \ie primitives, of individual manipulations.

\red{
Specific attention has also been directed to understanding action by means of hand-object interactions (many times: ``grasping", \citep{Elliott1984,Cutkosky1989,Ekvall2005,Feix2009,Wimmer2011,Bullock2013,Liu2016}). This issue is complex and can be quite confounding when trying to get closer to the semantics of actions. To be able to understand this problem we first need to better introduce our approach and we will therefore discuss the issue of hand-object interactions at greater length only later (in section~\ref{sec:discussion}).
}

\begin{table*}[!t]
\begin{center}
    \caption{\red{Comparison of recent action recognition approaches}} 
\scalebox{1.0}{
\begin{tabular}{ l  c  c  c  c  c  c  c  c  c  c  c c   p{6cm} }
    \\
    \multicolumn{1}{l}{Paper}   
        &  \mcrot{1}{l}{90}{\twoelementtable{Action}{Segmentation}} & \phantom{p}  
        &  \mcrot{1}{l}{90}{\twoelementtable{Parallel}{Actions}} & \phantom{p} 
        &  \mcrot{1}{l}{90}{\twoelementtable{Viewpoint}{Invariance}} & \phantom{p} 
        &  \mcrot{1}{l}{90}{Semantic} 
        &  \mcrot{1}{l}{90}{Motion} 
        &  \mcrot{1}{l}{90}{Multi-agent}
        &  \mcrot{1}{l}{90}{\twoelementtable{Object}{Information}} & \phantom{p} 
        &  \mcrot{1}{l}{90}{Depth Cue}
        &  \multicolumn{1}{c}{Comment}  \\
    \midrule \midrule
        \cite{Badler1975}
        & \multicolumn{2}{c}{-} & \multicolumn{2}{c}{-} & \multicolumn{2}{c}{\OK} & \OK & \OK & - & \multicolumn{2}{c}{\OK} & -
        &   Not applied to real image streams \\
        \hline
        \cite{Brand97}
        & \multicolumn{2}{c}{-} & \multicolumn{2}{c}{-} & \multicolumn{2}{c}{-} & \OK & \OK & - & \multicolumn{2}{c}{\OK} & -
        &   The event detection and scoring steps are hand-tuned, not adaptive or robust \\
        \hline
        \cite{Bobick1998}
        & \multicolumn{2}{c}{-} & \multicolumn{2}{c}{-} & \multicolumn{2}{c}{-} & \OK & \OK & - & \multicolumn{2}{c}{-} & -
        &   Considers simple hand gestures \\
        \hline
        \cite{Rui00}
        & \multicolumn{2}{c}{\OK} & \multicolumn{2}{c}{-} & \multicolumn{2}{c}{-} & - & \OK & - & \multicolumn{2}{c}{\OK} & -
        &   Highly depends on the observed motion pattern \\
        \hline
        \cite{Minnen2003}
        & \multicolumn{2}{c}{\OK} & \multicolumn{2}{c}{-} & \multicolumn{2}{c}{-} & \OK & - & - & \multicolumn{2}{c}{-} & -
        &   Not applicable to complex actions with cluttered scenes \\
        \hline
        \cite{Zhong2004}
        & \multicolumn{2}{c}{\OK} & \multicolumn{2}{c}{-} & \multicolumn{2}{c}{-} & - & - & \OK & \multicolumn{2}{c}{-} & -
        &   Depends on the predefined window size required for temporal action segmentation \\
        \hline
        \cite{STIP2005}
        & \multicolumn{2}{c}{-} & \multicolumn{2}{c}{-} & \multicolumn{2}{c}{-} & - & \OK & \OK & \multicolumn{2}{c}{-} & -
        &  Highly depends on the observed scene context \\
        \hline
        \cite{Ryoo2006}
        & \multicolumn{2}{c}{\OK} & \multicolumn{2}{c}{-} & \multicolumn{2}{c}{\OK} & \OK & - & \OK & \multicolumn{2}{c}{-} & -
        &  Poor performance when having  missing or noisy sub-events \\
        \hline
        \cite{Sridhar08}
        & \multicolumn{2}{c}{-} & \multicolumn{2}{c}{-} & \multicolumn{2}{c}{\OK} & \OK & - & \OK & \multicolumn{2}{c}{-} & -
        &  Large and complex activity graphs need to be decomposed \\
        \hline
        \cite{Gupta2009}
        & \multicolumn{2}{c}{\OK} & \multicolumn{2}{c}{-} & \multicolumn{2}{c}{-} & \OK & \OK & - & \multicolumn{2}{c}{\OK} & -
        &  Depends on the scene context (\eg object texture and pose) \\
        \hline
        \cite{Kjellstrom11}
        & \multicolumn{2}{c}{-} & \multicolumn{2}{c}{-} & \multicolumn{2}{c}{-} & \OK & \OK & - & \multicolumn{2}{c}{\OK} & -
        &  Object interactions are not considered \\
        \hline
        \cite{Wang2011}
        & \multicolumn{2}{c}{-} & \multicolumn{2}{c}{-} & \multicolumn{2}{c}{-} & - & \OK & \OK & \multicolumn{2}{c}{-} & -
        &   Highly depends on the scene context and  motion pattern \\
        \hline
        \cite{Fathi2011}
        & \multicolumn{2}{c}{\OK} & \multicolumn{2}{c}{-} & \multicolumn{2}{c}{-} & - & \OK & - & \multicolumn{2}{c}{\OK} & -
        &  Object recognition follows action identification \\
        \hline
        \cite{Yang13}
        & \multicolumn{2}{c}{-} & \multicolumn{2}{c}{-} & \multicolumn{2}{c}{-} & \OK & - & - & \multicolumn{2}{c}{\OK} & -
        &  Requires prior knowledge about  manipulated objects \\
        \hline
        \cite{Ramirez2013}
        & \multicolumn{2}{c}{-} & \multicolumn{2}{c}{-} & \multicolumn{2}{c}{\OK} & \OK & - & \OK & \multicolumn{2}{c}{\OK} & -
        &  Abstract approach but also employs object information \\
        \hline
        \cite{Wang2013}
        & \multicolumn{2}{c}{-} & \multicolumn{2}{c}{-} & \multicolumn{2}{c}{-} & - & \OK & \OK & \multicolumn{2}{c}{-} & -
        & Highly depends on the followed motion pattern \\
        \hline
        \cite{Koppula13}
        & \multicolumn{2}{c}{\OK} & \multicolumn{2}{c}{-} & \multicolumn{2}{c}{-} & \OK & \OK & - & \multicolumn{2}{c}{\OK} & \OK
        & Complex approach, employing human skeleton information, object segments and object tracks\\
        \hline
        \cite{Li_2015_CVPR}
        & \multicolumn{2}{c}{-} & \multicolumn{2}{c}{-} & \multicolumn{2}{c}{\OK} & - & \OK & - & \multicolumn{2}{c}{\OK} & -
        & Depends on the scene context and motion profiles\\
        \hline
        \cite{Wei2016}
        & \multicolumn{2}{c}{\OK} & \multicolumn{2}{c}{\OK} & \multicolumn{2}{c}{-} & \OK & - & - & \multicolumn{2}{c}{\OK} & \OK
        & Requires human skeleton and depends on the object recognition \\
        \hline
        \textbf{Ours}
        & \multicolumn{2}{c}{\OK} & \multicolumn{2}{c}{\OK} & \multicolumn{2}{c}{\OK} & \OK & - & \OK & \multicolumn{2}{c}{-} & \OK
        &  Requires multi-object and hand tracking  \\
        \hline
        
    \bottomrule
\end{tabular}
\label{table:soacomparison}}
\end{center}
\end{table*}


Recent works such as \cite{Koppula13} described a Markov random field based model for decomposing and  labeling the sequences of human sub-activities together with manipulated object roles. In the modeling process they employed human skeleton information, object segments and the observed object tracks. Likewise, \cite{Gupta2009} introduced a Bayesian model by using hand trajectories and hand-object interactions while segmenting and estimating observed manipulation sequences. In \cite{Fathi2011}  hierarchical models of manipulations were learned with weak supervision from an egocentric perspective without using depth information. In contrast to our framework, these approaches are not suitable for detecting and recognizing  parallel streams of actions since the applied models can only assign one label to each computed temporal segment.

Following this analysis we believe that the here presented work is the first study that applies semantic reasoning in order to decompose chained manipulation sequences and to recognize embedded serial and parallel (overlapping) manipulation streams in conjunction with the manipulated objects without employing any prior object knowledge.

\red{
In Table~\ref{table:soacomparison} we provide a detailed comparison of some recent action recognition approaches together with our proposed method.
This side-by-side comparison shows which approaches can perform joint action segmentation (second column), detect parallel actions (third column) and which are viewpoint invariant (fourth column). 
Those are three main features come with the proposed SEC framework.
The fifth column of the table indicates which of those methods employ high-level action semantics for the recognition process. Although there exist many different semantic approaches, to the best of our knowledge, the SEC framework is the only one, which is fully grounded at the signal (\ie pixel) level. 
This SEC feature leads to extraction of key events, \ie primitives, of individual manipulations and also to the clustering of objects according to their roles in an action.
The next column highlights whether the motion profile of objects is being incorporated during the recognition phase. This feature makes action recognition biased to the followed movement pattern (trajectory). This can harm the method's power for generalization, which is not the case in our framework.
The seventh column shows which methods can handle multi-agent action streams. This property indicates whether the recognition method can deal with cluttered scenes where more than one subject manipulates multiple objects.
The last two columns respectively indicate whether the corresponding method requires prior object information or whether it depends on depth information. 
Different from other approaches, SECs do not employ any object information in advance but require depth for a better performance.
}

\section{Method}
\label{sec:method}

\begin{figure*}[!b]
\centering
 \includegraphics[scale=0.65]{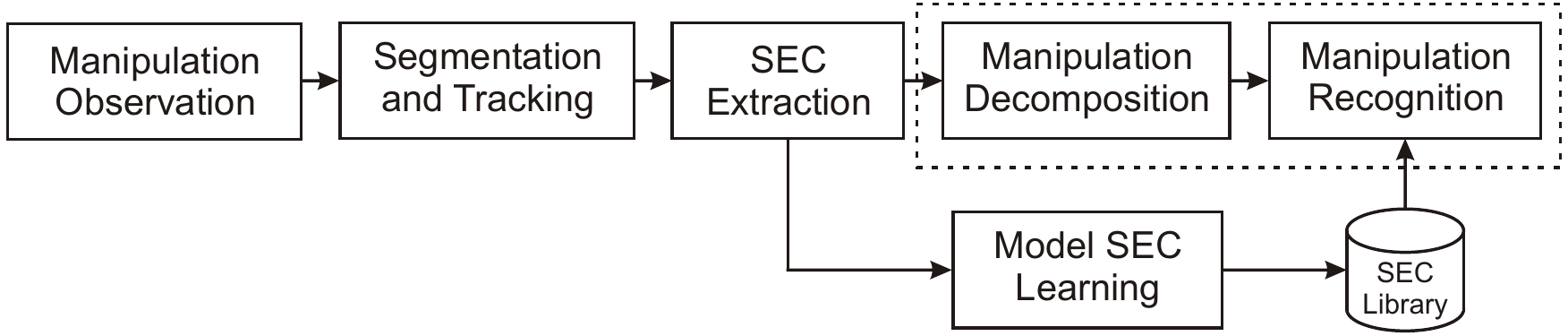}
  \caption{Block diagram of the algorithm. }
\label{fig:block_diagram}
\end{figure*}

Before describing the complete framework in detail, we will briefly provide an overview of each algorithmic step illustrated in Fig.~\ref{fig:block_diagram}.

  \begin{figure*}[!b]
       \centering
       \includegraphics[scale=0.5]{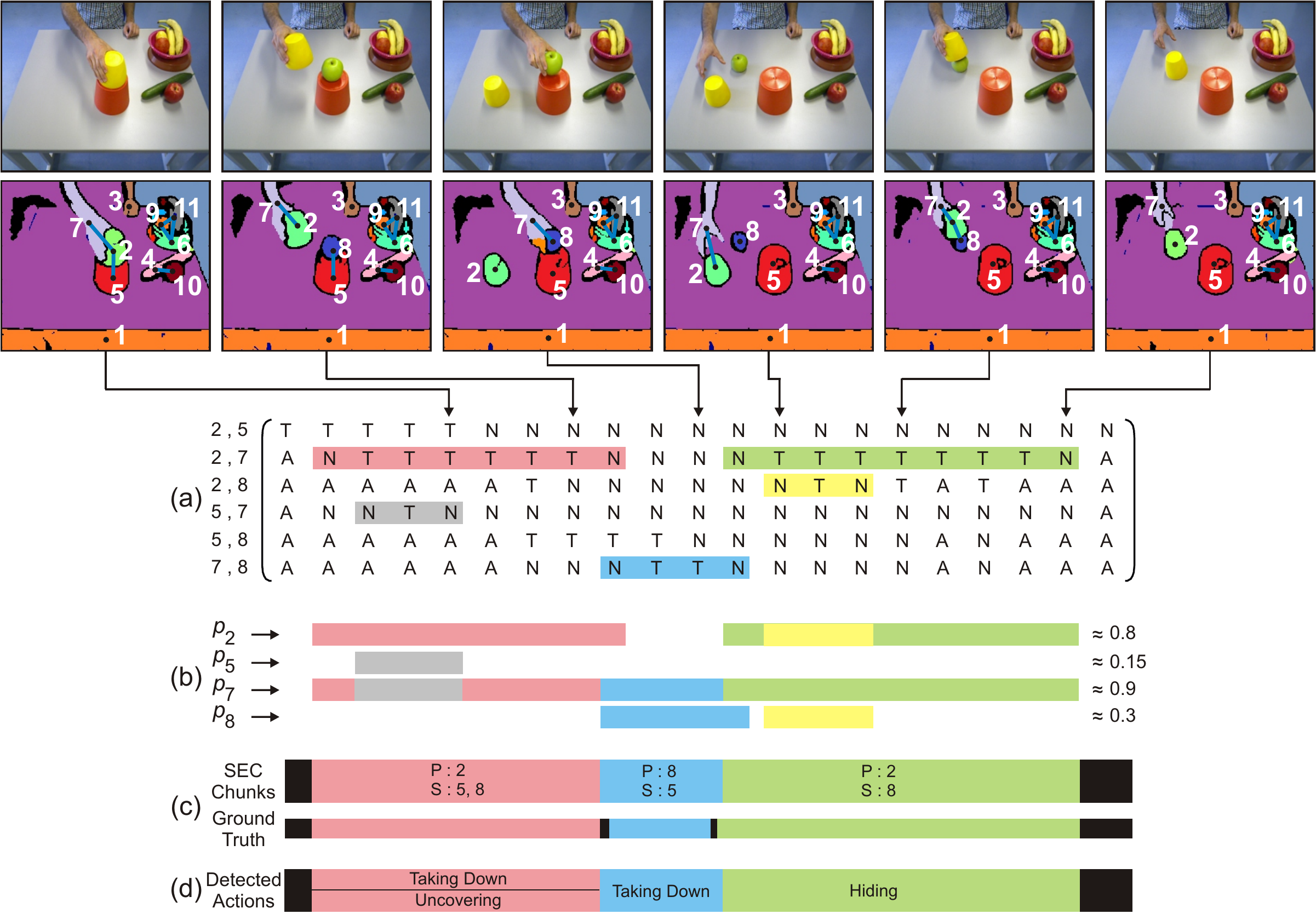}
        \caption{Semantic segmentation of a sample manipulation sequence where a hand is first replacing a bucket, putting an apple down and then hiding it  with the same bucket.
(a)~The extracted event chain where each column corresponds to one {\it key frame}, some of which are shown on the top with original images, respective objects (colored regions), and main graphs. Rows are spatial relations between object pairs, e.\,g. between the yellow ($2$) and red buckets ($5$) in the first row. Possible spatial relations are $N$, $T$, and $A$ standing for {\it Not touching}, {\it Touching}, and {\it Absence}. Each colored block in the SEC indicates a sequence of $[N, T, \cdots, T, N]$ relations.
(b)~Computed probabilities of each object to estimate the {\it manipulator}. For instance, the object number $2$ exists only in the first three rows of the SEC, therefore, detected blocks only in these rows are superimposed and assigned for that object to calculate the probabilities given on the right.
(c)~Decomposed SEC segments with respect to the ground truth. Black blocks represent null actions. {\it P} and {\it S}
stand for the estimated {\it primary} and {\it secondary objects}.
(d)~Detected manipulation types at each segment.}
\label{fig:samplechainedactionSEC}
\end{figure*}

The proposed semantic action temporal segmentation and recognition framework is triggered with the observation of manipulation actions demonstrated by a human. The image sequence of any observed manipulation is first segmented to separately track each object-like entity (including hand) in the scene by using computer vision methods \citep{Abramov10,AbramovRGBD12}. Note, explicit object information is not provided and the method just tracks ``image segments".  Tracked image segments, \ie objects, are then represented by scene graphs to derive a matrix like manipulation representation, the so-called Semantic Event Chain (SEC). Objects are the graph nodes and edges exist between objects that touch each other. Graphs are only stored when their topology changes (i.e. when nodes ore edges are formed or deleted). Hence, essentially we record the touching or un-touching events between objects here. The core algorithm to extract event chains has been described elsewhere \citep{Aksoy2011}. In the first step we create a SEC library of various atomic manipulations (e.g. {\it Cutting} or {\it Stirring}) by learning an event chain model for each individual type with  a method introduced in \cite{AksoyRAS2015}. This method is model-free and is only based on the intrinsic correlations in the topology changes of the graphs, which are highly characteristic for different atoms. As mentioned, these two steps (SEC algorithm and SEC learning) been described earlier \citep{Aksoy2011,AksoyRAS2015} and the main contribution of this paper lies elsewhere: It is the semantic segmentation and recognition of long and complex manipulation sequences, as depicted by a dashed box in Fig.~\ref{fig:block_diagram}. In brief: The event chain representation of the observed manipulation is first scanned to estimate the main manipulator, \ie the hand, in the scene without employing any object recognition method. Solely based on the interactions between the hand and manipulated objects in the scene, the event chain is decomposed into segments. Those are further fragmented into sub-units to detect parallel action streams. Each parsed SEC segment is then compared with the model SECs in the library to decide whether the current SEC sample belongs to one of the known manipulation models or represents a novel manipulation. The proposed framework is running in an automated and unsupervised manner to monitor chained manipulation sequences performed either sequentially or in parallel.

In the next sections we will present the core algorithmic components with all details, however, those which have been introduced elsewhere will only be briefly summarized.

\subsection{Manipulation Observation}
\label{sec:observation}

In this work, we address the automatic temporal segmentation and monitoring of long and complex manipulation sequences in which a human is manipulating multiple objects in various orders to perform specific tasks, such as ``{\it making a sandwich}" or ``{\it preparing a breakfast}". During the observation phase, the demonstrated manipulation is recorded from the subject's own point of view with a static $RGB-D$ camera since we are interested in the spatiotemporal interactions between the manipulated objects and hands. The top row in Fig.~\ref{fig:samplechainedactionSEC} depicts some original scene images from a sample chained manipulation demonstration.

\subsection{Segmentation and Tracking}
\label{sec:segmentationandtracking}

The image segmentation algorithm is based on the color and depth information fed from the Kinect device and uses phase-based optical flow \citep{Pauwels10} to track objects between consecutive frames. Data transmission between  different pre-processing sub-units is achieved with the modular system architecture described in \cite{PaponOcl2011}. Since segmentation and tracking approaches are not in the core of this paper and were comprehensively described  elsewhere \citep{Abramov10,AbramovRGBD12}, we omit details here.

Note that for image segments, \ie objects, we will always use more descriptive human terms like {\it hand}, {\it bucket}, etc., but we emphasize that the system has no such knowledge and it entirely lies on consistently tracked image segments.

\subsection{Semantic Event Chain (SEC) Extraction}
\label{sec:secs}

Each segmented image is represented by a graph: nodes represent object centers and edges indicate whether two objects touch each other or not. 
By using depth information we exclude the graph node for the background (supporting surface) since it is, in general, not employed as the main object manipulated in the action.
By using an exact graph matching technique, the framework discretizes the entire graph sequence into decisive main graphs. A new main graph is identified whenever a new node or edge is formed or an existing edge or node is deleted. Thus, each main graph represents a ``key frame" in the manipulation sequence, where a discrete change has happened. All issued main graphs form the core skeleton of the SEC, which is a matrix where rows are spatial relations (e.\,g. touching) between object pairs and columns describe the scene configuration at the time point when a new main graph has occurred.

Fig.~\ref{fig:samplechainedactionSEC}~(a) depicts the SEC representation for a sample chained manipulation demonstration, in which a hand is first replacing a bucket, then putting an apple down  on the table and then hiding it  with the same bucket. For instance, the first row of the SEC represents the spatial relations between graph nodes $2$ and $5$ which are the yellow and red buckets, respectively. On the top of Fig.~\ref{fig:samplechainedactionSEC}~(a) some sample {\it key frames} including original images, respective objects (colored regions), and corresponding main graphs are given to illustrate topological configurations at the related SEC columns.

Possible spatial relations are {\it Not touching (N)}, {\it Touching (T)}, and {\it Absence (A)}, where $N$ means that there is no edge between two objects, \ie graph nodes corresponding to two spatially separated objects, $T$ represents a touching event between two neighboring objects, and the absence of an object yields $A$. In the event chain representation, all pairs of objects need to be considered once, however, static rows which do not contain any change from $N$ to $T$ or vise versa are deleted as being irrelevant. For instance, the relation between the left and right hand is always $N$ and never switches to $T$ to trigger an event, therefore, the respective row is ignored in the event chain. Consequently, the SEC in Fig.~\ref{fig:samplechainedactionSEC}~(a) encodes relations only between objects $2,~5,~ 7,$ and $8$, although  many more objects are existing in the scene. Hence, the semantics of the manipulation is now represented by a $6 \times 21$ matrix despite of having approximately $1100$ frames in the entire demonstration. The SEC extraction explained briefly in this section has been described in detail in \cite{Aksoy2011}.

\subsection{Learning of Model SECs}
\label{sec:learning}

In this section, we will briefly describe both the learning method employed to explore model SECs for single atomic manipulations  and the semantic similarity measure between two event chains. We, however, omit the finer details here and refer the interested reader to \cite{AksoyRAS2015} for a comprehensive description of those approaches.

The main aim of the learning method is to generate a vocabulary of single atomic manipulations, \eg~{\it Putting},
{\it Hiding}, or {\it Pushing}. Such a vocabulary can then be employed to monitor the decomposed long manipulation sequences (see Fig.~\ref{fig:block_diagram}).

\begin{figure}[!t]
\centering
 \includegraphics[scale=0.5]{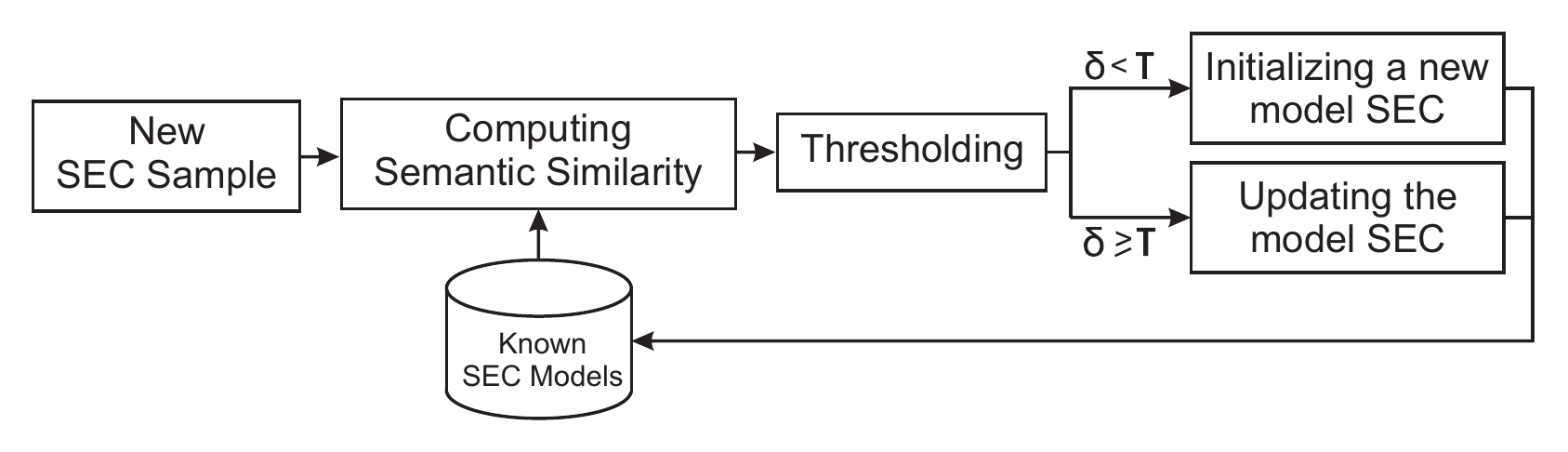}
  \caption{Overview of the learning framework.}
\label{fig:learning_overview}
\end{figure}

The learning approach essentially searches for common spatiotemporal information embedded in the rows and columns of event chains derived from a training manipulation set. Fig.~\ref{fig:learning_overview} shows the overview of the learning approach. A new model is initiated with the first SEC sample of an unknown atomic manipulation. Once the next demonstration is observed, the respective SEC is derived and compared with the already known SEC models. If the semantic similarity ($\delta$) between this novel SEC sample and any of the known models is higher than a threshold ($\tau$), the corresponding model is updated with the new sample. Otherwise, the SEC sample is labeled as a new model. The threshold value $\tau$ is directly estimated  from the distribution of semantic similarities between observed SEC sample and those known models.
To update an existing model, the learning procedure just needs to search for all common rows and columns existing both in the new SEC sample and the model. In the case of having additional rows or columns in the new SEC, the model is extended by these extra ones. Finally, the model SEC consists of only those rows and columns observed frequently in the new acquired SEC samples.  The learning framework works in an on-line unsupervised manner as described in detail in \cite{AksoyRAS2015}. A batch mode implementation had already been introduced in \cite{Aksoy2011}.

In order to measure the semantic similarity between two event chains, we basically compare rows and columns of SECs using simple sub-string search and counting algorithms. Relational changes are considered while comparing the rows, whereas for the columns the temporal order counts. We first search for the correspondences between rows of two event chains since rows can be shuffled. The searching process compares and counts equal entries of one row against other rows using a standard substring search which does not rely on dimensions and allows comparing arbitrarily long manipulation actions. We then examine the order of columns to get the final similarity result. Details for similarity calculations are given in \cite{AksoyRAS2015}.

\begin{figure}[!t]
\centering 
 \includegraphics[trim={3cm 3cm 5cm 0.65cm},scale=0.45]{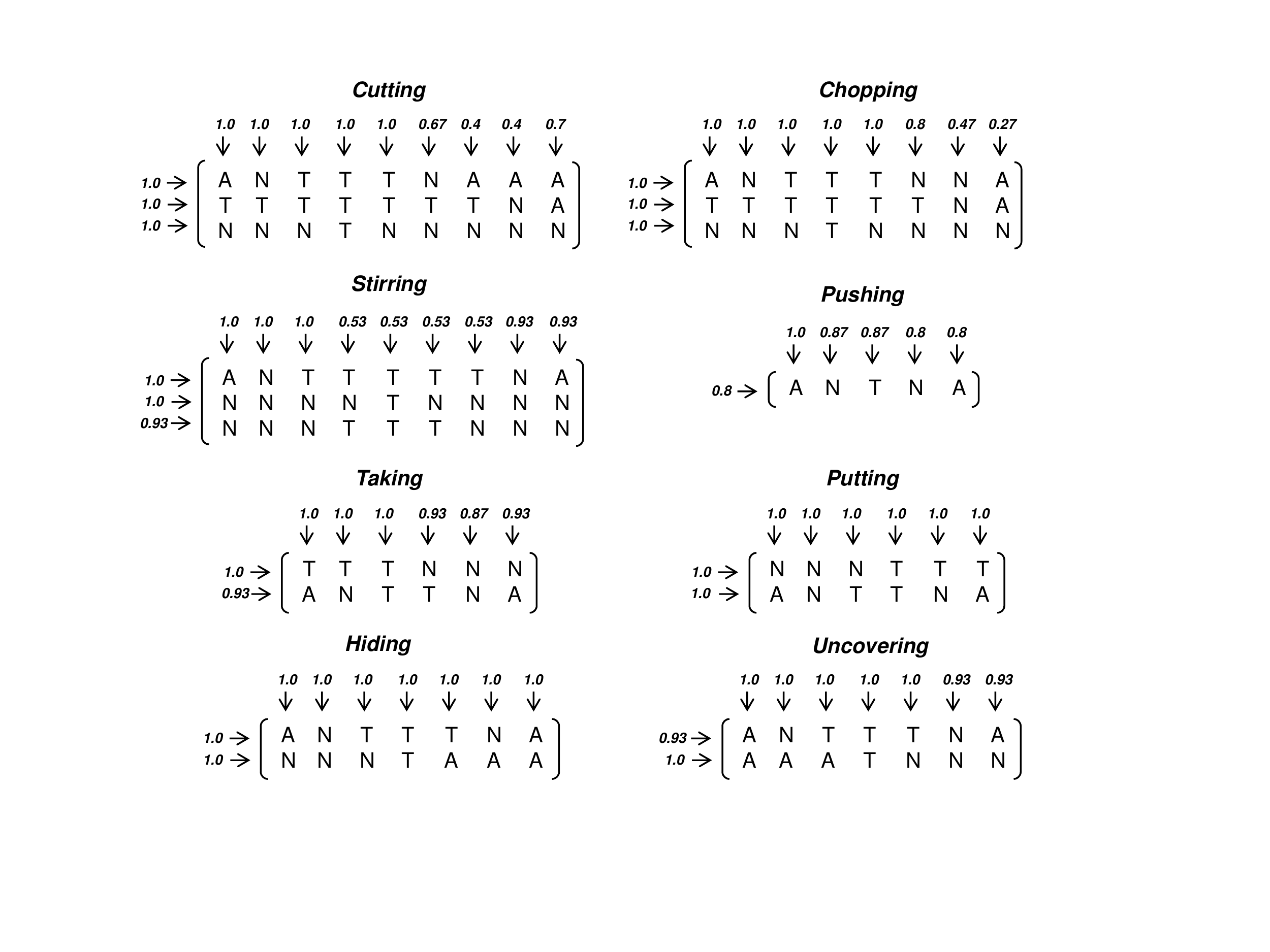}
  \caption{Complete learned SEC models for eight different manipulations. Weight values shown with arrows on the left and top respectively indicate how often each row and column in the SEC is obtained in the trained samples. }
\label{fig:learned_sec_models}
\end{figure}

Fig.~\ref{fig:learned_sec_models} shows learned SEC models for eight different atomic manipulations in the ManiAc dataset \citep{AksoyRAS2015} explained in Section~\ref{sec:maniacdataset}. Each arrow on the top of SEC models indicates the weight values of these most commonly observed event chain columns. Weight values depicted with arrows on the left represent  how often each SEC row is obtained in the trained samples. It can be seen that in all models rows are quite commonly observed in the trained samples as their weight values are close to $1$. This is, however, not the case for a few SEC model columns. For instance, the weight value for the last column of the \textit{Chopping} model drops to $0.27$. This is because even though each subject grasps a tool and chops an object in a similar temporal order, they leave the scene in different orders; for example, one subject first removes the hand supporting the object to be chopped and then withdraws the hand holding the tool whereas another subject either does it the other way around or removes both hands at the same time. Another reason of having smaller weight values is the noise propagated from the segmentation and tracking components as observed in the \textit{Cutting} model. Nevertheless, we can extract all these variations that occurred due to the nature of manipulation or noise and pick the most often observed states as a representative model for each manipulation action.

It is obvious that these $N$-$T$-$A$ patterns in the learned SEC models are very unique, except for \textit{Cutting} and \textit{Chopping} which have a quite similar SEC structure. This is because both manipulations semantically represent the same manipulation consequence, hence, both have the same fundamental action primitives, i.e. similar columns in the event chains. The only differences are mostly in the followed trajectories and velocity of the movements which are not captured by SECs. We discuss about such naturally emerging high semantic similarities in Section~\ref{sec:maniacdataset}.

It is also important to highlight that some SEC models have symmetric patterns, such as those in \textit{Hiding} and \textit{Uncovering} or \textit{Taking} and \textit{Putting} models. This is fundamentally very correct since backward playing of any of these manipulations will lead to its symmetrical counterpart. This is a very important feature coming with the semantic event chain representation of manipulations.

\subsection{Manipulation Temporal Segmentation}
\label{sec:manipulationdecomposition}

Once the SEC pattern of a manipulation sequence is derived, we continue with the temporal segmentation phase which considers the semantic information embedded in the event chain. The temporal segmentation method first searches for an object which plays the main role in the manipulation, or in other words, which acts as a {\it manipulator}. We assume that each manipulation is driven by one such main actor, \eg~a {\it hand} and that it is most frequently interacting with the objects in the scene. To make the rest of the algorithm more clear, we start with the assumption of having only {\it single-hand} manipulations, however, this can be extended to multiple hands as will be discussed in section~\ref{sec:mot_dataset}. For this framework we employ the following reasonable action descriptive rules:

\begin{itemize}

\item

The {\it manipulator} can purposefully manipulate, \ie~{\it touch}, only one object at a time, which will be named  {\it primary object}, \eg~a {\it knife}.

\item
The manipulation sequence can consist of multiple {\it primary objects}. Each, however, has to be separately manipulated at different time intervals. For instance, the hand cutting a cucumber with a knife is not allowed to stir milk with a spoon unless releasing the knife first.

\item
All other objects interacting with the {\it primary objects} will be called  {\it secondary objects}, \eg the cucumber to be cut.
\end{itemize}

Those rules, first introduced in \cite{Woergoetter2013}, form the main skeleton of our proposed temporal segmentation method.

We first start converting these rules into meaningful spatial relational sequences to make them compatible with the SEC representation. For instance, these rules require that the event chain must have at least a row holding spatial relations between the {\it manipulator} and {\it primary object} defined as:

\begin{equation}
{\it manipulator}, {\it primary~object} ~~~
\begin{bmatrix}
   N & T & \cdots & T & N
\end{bmatrix},
\label{ntn_sequence}
\end{equation}

where the {\it manipulator} is first not touching (N) the {\it primary object}, then touches (T) {\it primary object} to apply a certain task on it. Depending on the manipulation, the temporal length of the touching (T) relation can vary. Finally, the {\it manipulator} releases (N) the {\it primary object} and continues with a different  {\it primary object}.

We note that as there is no object recognition method existing to identify graph nodes, we first need to identify the {\it manipulator} or {\it primary/secondary objects} using just the naked graph nodes, based only on the above introduced action descriptive rules.

\subsubsection{Estimating the Manipulator}
\label{sec:extractingmanipulator}

To achieve this, we apply probabilistic reasoning to estimate object roles in the manipulation. Probability values for each object are assigned based on similarities of their relations with Eq.~\eqref{ntn_sequence} and the length of their touching relations.

Let $\xi$ be a semantic event chain with the size of $n \times m$ and assume that  $\xi$ includes $q$ different objects, the set of which can be written as

\begin{eqnarray}
\mathcal{S} =\{s_{1}, s_{2}, \cdots , s_{q}\}
\mathpunkt
\end{eqnarray}

The event chain $\xi$ can then be described as:

\begin{equation}
\xi =
\begin{bmatrix}
  s_{1,1},s_{1,2}\\
  s_{2,1},s_{2,2}\\
  \vdots \\
  s_{n,1},s_{n,2}\\
  \end{bmatrix}
=
\begin{bmatrix}
  r_{1,1} & r_{1,2} & \cdots & r_{1,m} \\
  r_{2,1} & r_{2,2} & \cdots & r_{2,m} \\
  \vdots  & \vdots  & \ddots & \vdots  \\
  r_{n,1} & r_{n,2} & \cdots & r_{n,m}
\end{bmatrix},
\label{sec_matrix}
\end{equation}

where $s_{i,:} \in \mathcal{S}$, and $r_{i,j} \in \{ A, N, T\}$ is representing the spatial relation between an object pair $s_{i,1}$ and $s_{i,2}$ at time $j$.

We assign a probability value $p_{k}$ to each object $s_{k}$ existing in  $\xi$ to define the likelihood of being the
{\it manipulator} as

\begin{eqnarray}
 \mathcal{P} = \{ p_{k}: \ k  \in [ 1,\cdots ,q  ] \} \
\mathkomma
\end{eqnarray}

\begin{eqnarray}
p_{k}= \frac{\sum^{m}_{j=1} \delta_{k,j}}{m}
\label{prob_segments}
\mathkomma
\end{eqnarray}

\begin{eqnarray}
\resizebox{.8\hsize}{!}{$\delta_{k,j} = \left\{
\begin{array}{l l}
  1 & \quad \text{if ~ $s_{k} \in s_{i,:}$ ~ , ~ $[N, T, \cdots, T, N] \in r_{i,:}$   } \\
    & \quad \text{ and ~ $r_{i,j} \in [N, T, \cdots, T, N]$ , $i  \in [ 1,\cdots ,n  ] $ }\\
  0 & \quad \text{else}\\
\end{array}
\right.
\mathkomma
$
}
\end{eqnarray}

where $\delta$ essentially investigates how wide a touching event $T$ expands over all the temporal length of $\xi$ in the case of having the relational sequence of $[N, T, \cdots, T, N]$ (given in Eq.~\eqref{ntn_sequence}) in all rows (\ie $s_{i,:}$) that include the respective object $s_{k}$. The {\it manipulator} is finally estimated as the object $s_{k^{\ast}}$ with the highest probability; that is,
	
\begin{equation}
k^{\ast}=   \underset{1 \leq k \leq q}{\operatorname{arg\,max} }~( p_{k})
\label{prob_manipulator}
\mathpunkt
\end{equation}

In Appendix~\ref{sec:manipulatorestimationappendix} we provide the pseudocode for estimating the {\it manipulator} from Eqs.~\eqref{ntn_sequence}$-$\eqref{prob_manipulator}.

For example, the colored blocks given in the SEC in Fig.~\ref{fig:samplechainedactionSEC}~(a) indicate where   sequences of $[N, T, \cdots, T, N]$, similar to the one given in Eq.~\eqref{ntn_sequence}, are detected. Fig.~\ref{fig:samplechainedactionSEC}~(b) links those blocks to the corresponding objects in the SEC to indicate which object has the longest block, \ie highest probability value. For instance, as object number $2$ exists only in the first three rows of the SEC, detected blocks only in these rows will be superimposed and assigned to that object. On the right side of Fig.~\ref{fig:samplechainedactionSEC}~(b), the final probability values computed from Eq.~\eqref{prob_segments} 
are given. Since  blocks associated with object number $7$ cover the widest temporal stretch along the SEC, it is correctly estimated as the {\it manipulator} from Eq.~\eqref{prob_manipulator}.

\subsubsection{Decomposing SECs}
\label{sec:decomposingsecs}

Following the estimation of the {\it manipulator}, the SEC is ready to be decomposed into shorter segments. The temporal segmentation proceeds considering the $[N, T, \cdots, T, N]$ sequences that belong to the {\it manipulator}, since any change from $N$ to $T$ and from $T$ to $N$ defines the natural start and end points of the manipulation. It is here important to note that we cannot directly assume each $[N, T, \cdots, T, N]$ sequence as a segment due to spurious spatial relations propagated from noisy segmentation and tracking phases. Therefore, we first apply a low pass filter to those rows with the {\it manipulator} and then label each time interval between an $[N, T]$ and $[T, N]$ change as a potential action segment. Each segment is assigned  a confidence value indicating the frequency of the touching relation. Finally, action segments that are encapsulated by others or share a common temporal zone are  merged to converge to the ultimate temporal segmentation of the manipulation.

Let $\mathcal{A}$ be a set of action segment candidates as

\begin{eqnarray}
\mathcal{A} =\{a_{1}, a_{2}, \cdots , a_{l}\}
\mathkomma
\end{eqnarray}

where each segment $a_{\phi}$ represents the area between the start and end time points of a  $[N, T, \cdots, T, N]$ sequence detected in each row $i$ that includes the {\it manipulator}; written as

\begin{eqnarray}
a_{\phi}=[t_{i}^{Start} ~~ t_{i}^{End} )
\mathkomma
\end{eqnarray}

where $t_{i}^{Start}$ , $t_{i}^{End}  \in [ 1,\cdots ,m ] $ and $m$ is the column number of $\xi$ in Eq.~\eqref{sec_matrix}.
Due to early vision problems, such as illumination variation or occlusion, noisy flickering spatial relations can occur at any action segment, $a_{\phi}$. We, thus, measure the rate of $T$ relations in each segment as a confidence value, $c_{\phi}$; that is computed as

\begin{eqnarray}
c_{\phi}= \frac{\sum^{t_{i}^{End}}_{j=t_{i}^{Start}} \theta_{i,j}}{t_{i}^{End} - t_{i}^{Start} }
\label{conf_val}
\mathkomma
\end{eqnarray}

\begin{eqnarray}
\theta_{i,j} = \left\{
\begin{array}{l l}
  1 & \quad \text{if ~ $r_{i,j}=T$  }\\
  0 & \quad \text{else}\\
\end{array} \right.
\mathkomma
\end{eqnarray}

\begin{figure}[!b]
       \centering
       \includegraphics[scale=0.6]{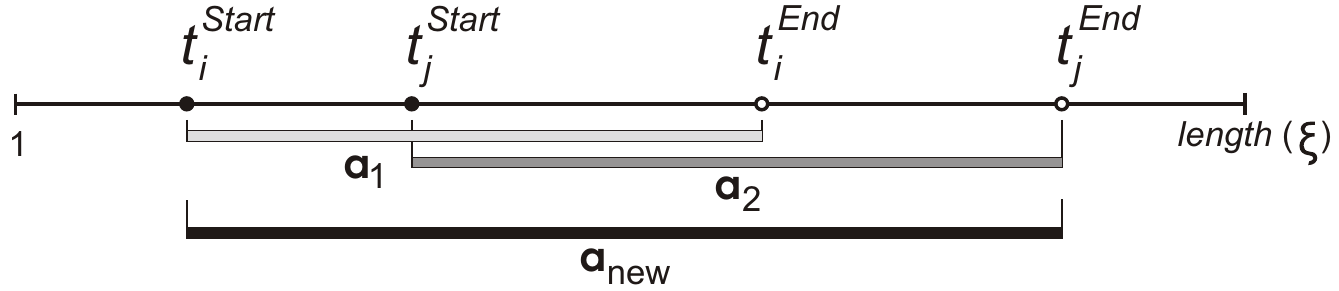}
        \caption{Merging action segments that share a common temporal field. }
\label{fig:segmentmerging}
\end{figure}

where $r_{i,j}$ is representing the spatial relation of the {\it manipulator} in $\xi$ in Eq.~\eqref{sec_matrix}. We then consider the action segments with higher confidence value than a predefined threshold $\tau_{conf}$. The confident segments are further compared to ignore those that are completely covered by others. We also merge segments that share a common field more than a threshold $\tau_{merge}$ (see Fig.~\ref{fig:segmentmerging}). Let assume that $a_{1}=[t_{i}^{Start} ~~ t_{i}^{End} )$  and $a_{2}=[t_{j}^{Start} ~~ t_{j}^{End} )$ are two segments as given in Fig.~\ref{fig:segmentmerging}.  In the case of having $\frac{|a_{1} 	\cap a_{2}|}{min(|a_{1}|,|a_{2}|)} \geq \tau_{merge}$, those two segments will be fused yielding a new segment $a_{new}$ with the length of $[t_{i}^{Start} ~~ t_{j}^{End} )$ as illustrated in Fig.~\ref{fig:segmentmerging}.
Note that $\tau_{conf}$ and $\tau_{merge}$ are chosen as $0.6$ in all our experiments.

If we come back to the example in Fig.~\ref{fig:samplechainedactionSEC}~(b), we see four candidate action fragments which are the red, blue, green, and gray blocks of the {\it manipulator}, \ie object number $7$. However, the gray block is ignored as it is entirely surrounded by the red one. Thus, the remaining three blocks construct the ultimate temporal points at which the manipulation will be cut. Fig.~\ref{fig:samplechainedactionSEC}~(c) illustrates the final temporal segmentation results together with the ground truth defined by a human. Note that the end point of each block in Fig.~\ref{fig:samplechainedactionSEC}~(c) is considered as the beginning of the next consecutive one. Compared to the ground truth, the frame-wise temporal segmentation accuracy of the manipulation sequence in Fig.~\ref{fig:samplechainedactionSEC}~(c) is computed as $96\%$.

\subsection{Manipulation Recognition}
\label{sec:manipulationrecognition}

In the recognition phase, we aim at identifying the types of performed manipulations for each decomposed SEC segment. The recognition process is based on the semantic similarities between the currently decomposed SEC segments and the pre-learned model SECs (see Section~\ref{sec:learning}).

Once the entire event chain is decomposed into smaller units, we  first distinguish the {\it primary} and {\it secondary objects} manipulated in each parsed segment. Recalling the action descriptive rules introduced in section~\ref{sec:manipulationdecomposition}, we define the object that is mostly interacting with the {\it manipulator} as the {\it primary object} and all other objects interacting with the {\it primary object} as the {\it secondary objects}. For instance, the event chain decomposed in Fig.~\ref{fig:samplechainedactionSEC}~(c) has three main pieces as indicated by red, blue, and green blocks, respectively. In the temporal interval of the red block (between the second and eighth columns of the SEC), the object number $2$ (the yellow bucket) is estimated as the {\it primary object} since it has  most touching events with the previously detected {\it manipulator}, \ie object number $7$. Next, objects $5$ and $8$ (the apple and red bucket) are found to be {\it secondary objects} because they are the only objects sharing a touching relation with the {\it primary object} within this same temporal interval. All estimated {\it primary} and  {\it secondary objects} in each parsed SEC segment are indicated in Fig.~\ref{fig:samplechainedactionSEC}~(c).

The main reason for reformulating the manipulations in terms of interactions between the {\it manipulator}, {\it primary} and {\it secondary objects} is two fold: First, we attempt to reduce the degree of noise in the decomposed event chain segments. As the action rules described in section~\ref{sec:manipulationdecomposition} do not allow the {\it manipulator} to interact with any object other than the {\it primary object}, we omit, for instance, the fourth row of the SEC in Fig.~\ref{fig:samplechainedactionSEC}~(a). This is because the {\it manipulator} (object number $7$) is accidentally touching the red bucket (object number 8) as highlighted by the gray block.
Details of such a high level de-noising process were described in \cite{AksoyRAS2015} to efficiently cope with noisy spatiotemporal information coming from the early vision stage.

Second, using this assumption we can also diagnose parallel streams of simultaneous manipulations by considering the fact that each manipulation has to have a unique {\it secondary object}. In other words, detection of multiple {\it secondary objects} may imply either noisy elements in the event chain or the existence of parallel manipulations. Hence, we treat each combination of the {\it manipulator}, {\it primary} and {\it secondary objects} as a separate manipulation hypothesis and choose the one, that has the highest semantic similarity with the learned SEC models, as the final recognition result.

For this we introduce a brute force combinatorial process which considers all combinations of the entire estimated {\it secondary objects} (which are only a few) together with the {\it manipulator} and {\it primary object} to accurately identify the actual performed manipulations.
The total number of the combinations can be computed as

\begin{eqnarray}
\mathcal{C}= \sum^{N}_{k=1} \frac{N!}{k!(N-k)!}
\label{combinatorial_hypotheses}
\mathkomma
\end{eqnarray}

where $N$ is the number of the existing {\it secondary objects} at a given SEC segment. We then use those combinations to generate various hypotheses that correspond to sets of manipulations. For instance, a hypothesis can represent either a single manipulation (\eg~ {\it Hiding}) or concurrently performed more manipulations such as {\it Putting} and {\it Pushing}. The crucial rule here is that each hypothesis must consist of  the  entire {\it secondary object} set.

\begin{figure}[!t]
       \centering
       \includegraphics[scale=0.6]{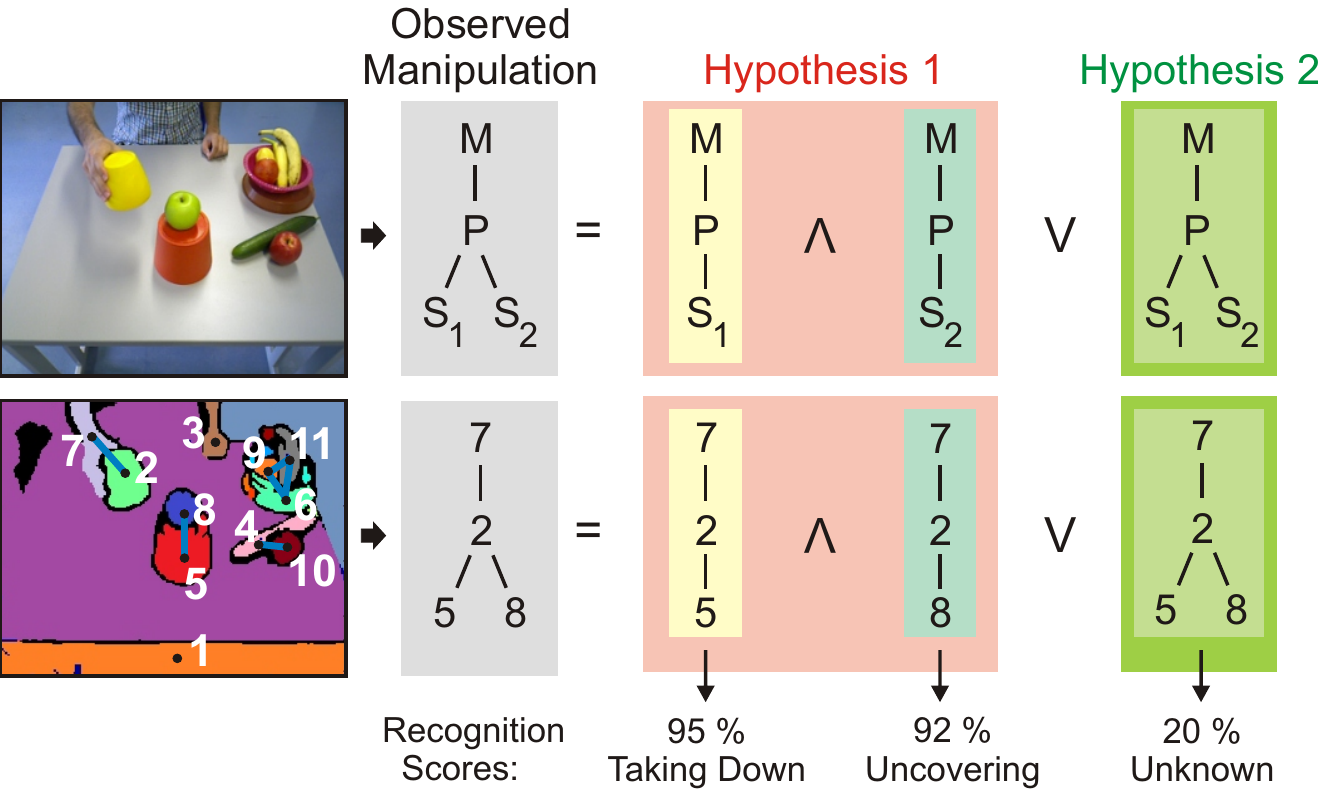}
        \caption{Detection of parallel manipulation streams in a decomposed SEC segment that has two {\it secondary objects}. {\it M}, {\it P}, and {\it S} stand for the {\it manipulator}, {\it primary} and {\it secondary objects}, respectively. As in the example of the SEC segment depicted in the red block in Fig.~\ref{fig:samplechainedactionSEC}~(c) (a sample key frame is also given here on the left), there are two possible hypotheses and each defines a different set of manipulation streams which are depicted by unique colors. Recognition score of each stream is given below.}
\label{fig:sec_hypothesis}
\end{figure}

\begin{figure*}[!t]
       \centering
       \includegraphics[scale=0.65]{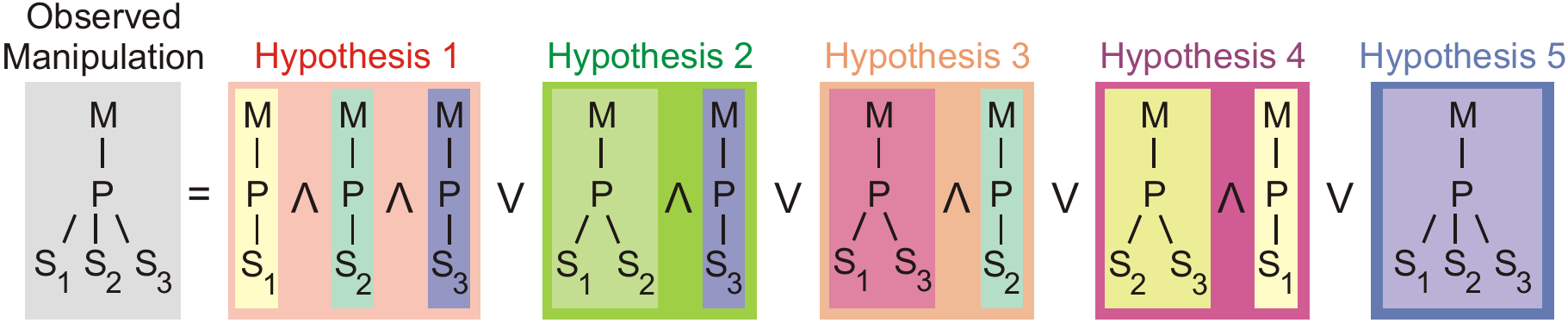}
        \caption{The entire hypothesis set estimated in the case of having three {\it secondary objects}. {\it M}, {\it P}, and {\it S} stand for the {\it manipulator}, {\it primary} and {\it secondary objects}, respectively.}
\label{fig:five_sec_hypotheses}
\end{figure*}

For instance, the first parsed SEC segment, depicted by the red block in Fig.~\ref{fig:samplechainedactionSEC}~(c), has two {\it secondary objects} (object numbers $5$ and $8$). Fig.~\ref{fig:sec_hypothesis} illustrates the computed two hypotheses  each of which has a different object combination, \ie manipulation stream. The first hypothesis is composed of two separate (parallel) manipulation streams, each utilizes one of the {\it secondary objects} as indicated by unique colors, whereas the next hypothesis employs  both {\it secondary objects} together as one manipulation stream. Unlike the {\it secondary objects}, in both hypotheses the {\it manipulator} and {\it primary object} are remaining the same. Note that, even though the scene involves many more objects, the number of hypotheses is remaining small due to the consideration of only those objects that are sharing touching events with the {\it primary object}. Thus, our approach does not suffer from  combinatorial explosion. The maximum number of combinations, \ie $\mathcal{C}$, observed so far in all our experiments, is $15$.
The entire hypothesis set, in the case of having three {\it secondary objects}, is depicted in Fig.~\ref{fig:five_sec_hypotheses}.

Next, we extract and smooth the corresponding event chains of all possible streams in each hypothesis and compare them semantically with the SEC models of atomic manipulations that have previously been learned and stored in the library. In the semantic comparison process (see section~\ref{sec:learning}), we introduce a similarity threshold $\tau_{sem}$ to explore whether two event chains belong to the same manipulation type. In the case of having a too low semantic similarity with any of the known SEC models, the respective event chain will then be assigned to {\it Unknown}. Unless otherwise stated, we keep the threshold value $\tau_{sem}$ as $72\%$ which has been found in an on-line unsupervised manner as introduced in \citep{AksoyRAS2015}.

The manipulation recognition task is now nothing else than computing the best hypothesis that has the highest semantic similarity with those individual models in the library. Note that if a hypothesis has multiple streams, the mean similarity value is considered to compare it with that of the other hypotheses. Fig.~\ref{fig:sec_hypothesis} shows at the bottom the final similarity scores in each hypothesis. As the first hypothesis has much higher recognition rate, our proposed approach successfully returns two parallel manipulation streams;
{\it Taking~Down} the yellow bucket (object 2) from the red bucket (object 5) while  {\it Uncovering} the green apple (object 8).
Fig.~\ref{fig:samplechainedactionSEC}~(d) illustrates the final manipulation types recognized at each decomposed SEC segment for the manipulation sequence depicted in Fig.~\ref{fig:samplechainedactionSEC}~(a).

\section{Results}
\label{sec:results}

In this section, we will provide experimental results from our proposed temporal segmentation and recognition method using different datasets. We first start benchmarking with our large Manipulation Action (ManiAc) dataset \citep{AksoyRAS2015} and will then continue with a recently published Manipulation Action Consequences (MAC) dataset \citep{Yang13}. The next section is covering the two-hand manipulations on the Multiple Object Tracking (MOT) dataset \citep{Koo14}. In the later section we will show baseline experiments conducted on these three datasets. \red{ In the very last section we will apply our SEC-based method to the MPII 
Cooking Activities dataset \citep{Rohrbach12} which involves long and parallel actions as RGB only image streams.}

\begin{figure}[!t]
 \centering
        \centering
      \includegraphics[scale=0.45]{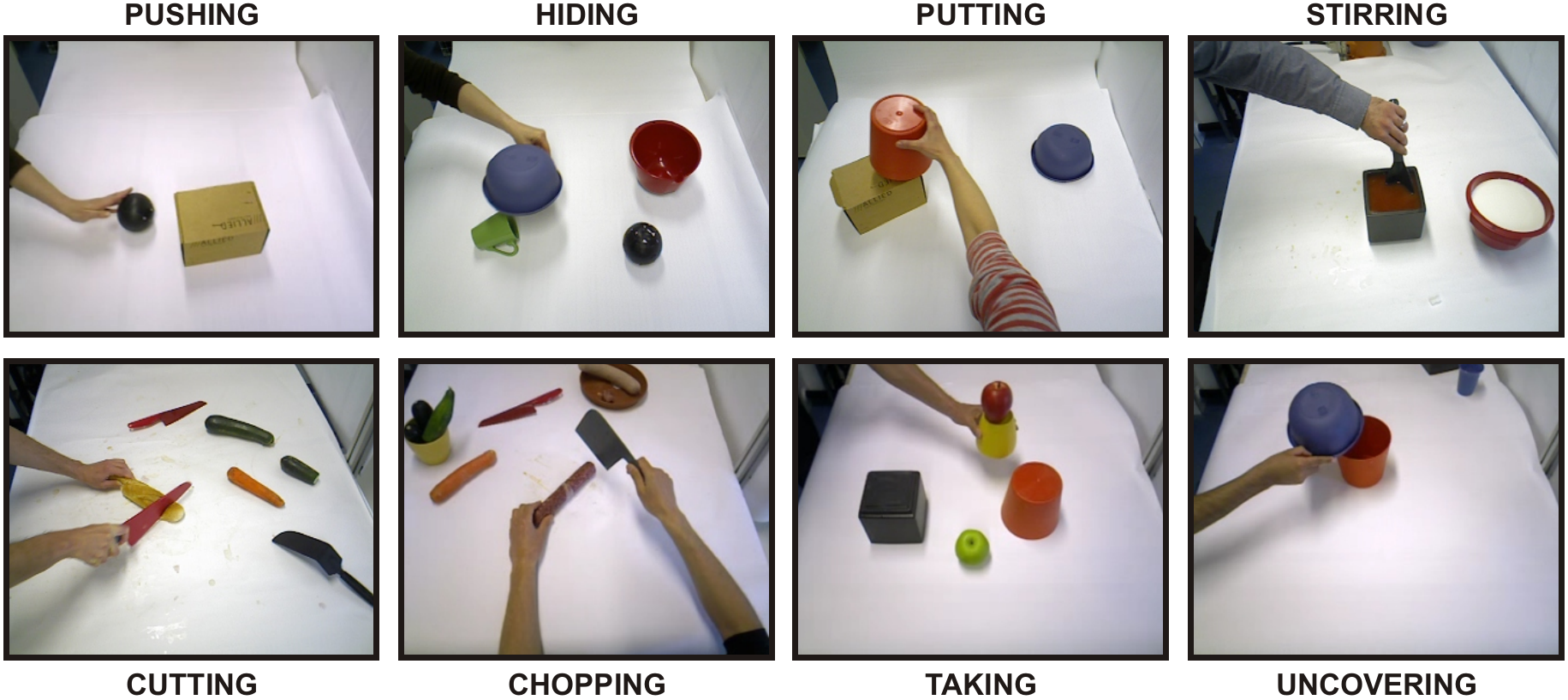}
      \caption{The ManiAc dataset with eight different single manipulation scenarios:  {\it Pushing}, {\it Hiding}, {\it Putting}, {\it Stirring}, {\it Cutting}, {\it Chopping}, {\it Taking}, and {\it Uncovering}.}
      \label{fig:trainingactions}

 \centering
         \includegraphics[scale=0.6]{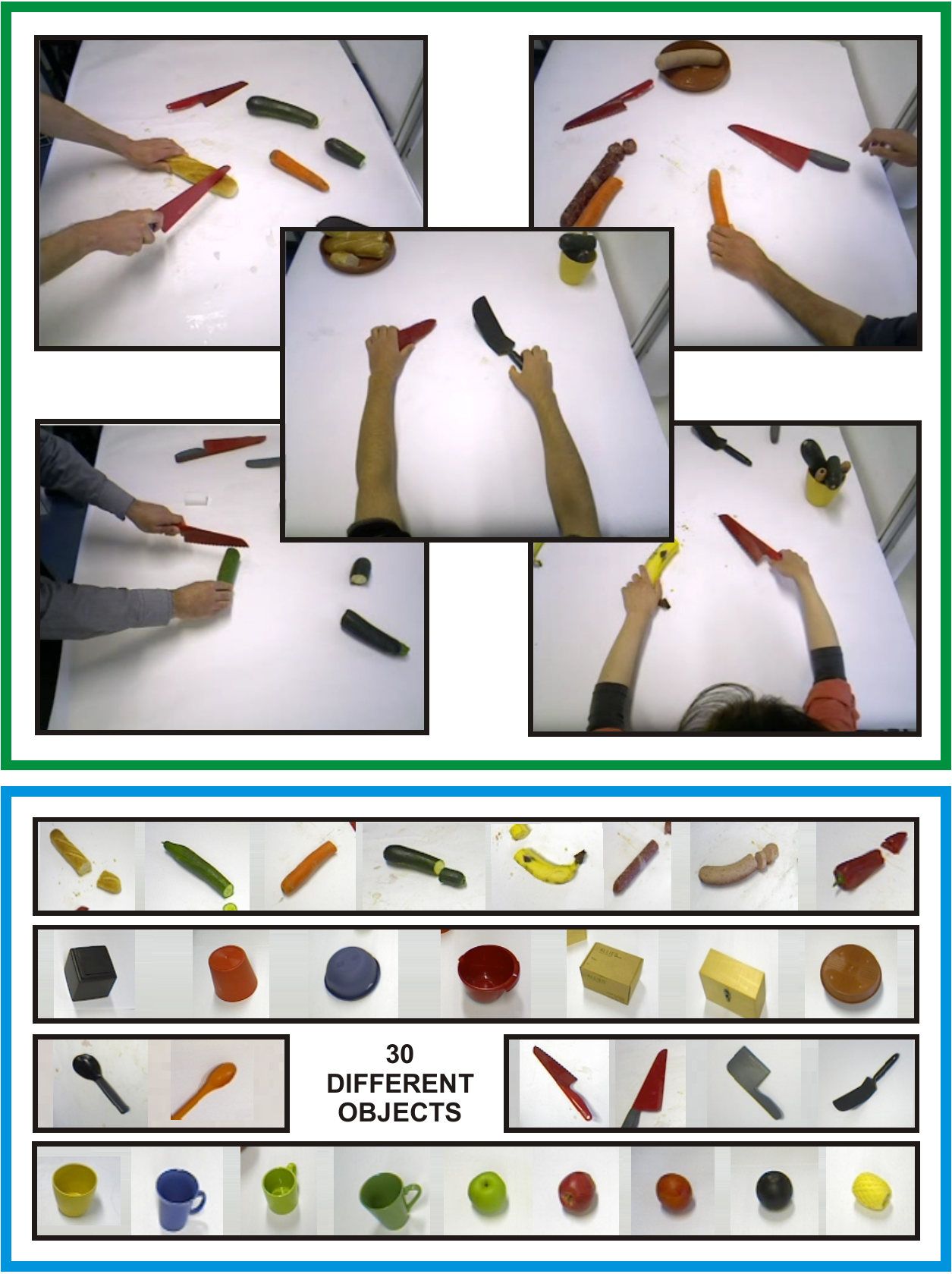}
      \caption{Sample images from the ManiAc dataset. In the green frame, a sample image from each demonstration of the {\it Cutting} action, performed by 5 different individuals, is given. The blue frame depicts 30 different objects manipulated in all 120 manipulation demonstrations.}
      \label{fig:training_cutting_versions}
\end{figure}
\subsection{Manipulation Action (ManiAc) Dataset}
\label{sec:maniacdataset}

The ManiAc dataset, introduced in our previous work \citep{AksoyRAS2015}, investigates eight different single atomic manipulation actions:  {\it Pushing}, {\it Hiding}, {\it Putting}, {\it Stirring}, {\it Cutting}, {\it Chopping}, {\it Taking}, and {\it Uncovering}. The complete dataset is publicly available at \url{www.dpi.physik.uni-goettingen.de/~eaksoye/MANIAC_DATASET}. Fig.~\ref{fig:trainingactions} shows sample frames from each action type. In the dataset, each manipulation has 15 different versions demonstrated by 5 different individuals using in total 30 various objects.   Fig.~\ref{fig:training_cutting_versions} depicts all objects used in the  dataset and some sample frames from the {\it Cutting} action demonstrated by $5$ different subjects. We benefit from this manipulation action dataset that consists of in total 120 single demonstrations in order to create a vocabulary of atomic manipulations by learning the semantics of actions, \ie the model SECs as described in section~\ref{sec:learning}. For this, all $15$ demonstrations of each of the $8$ single atomic manipulations were employed in a batch mode to derive a SEC model for each manipulation type. The learned $8$ models were then stored in the SEC library to be further used in the monitoring stage.

\begin{figure}[!b]
\centering
 \includegraphics[scale=0.5]{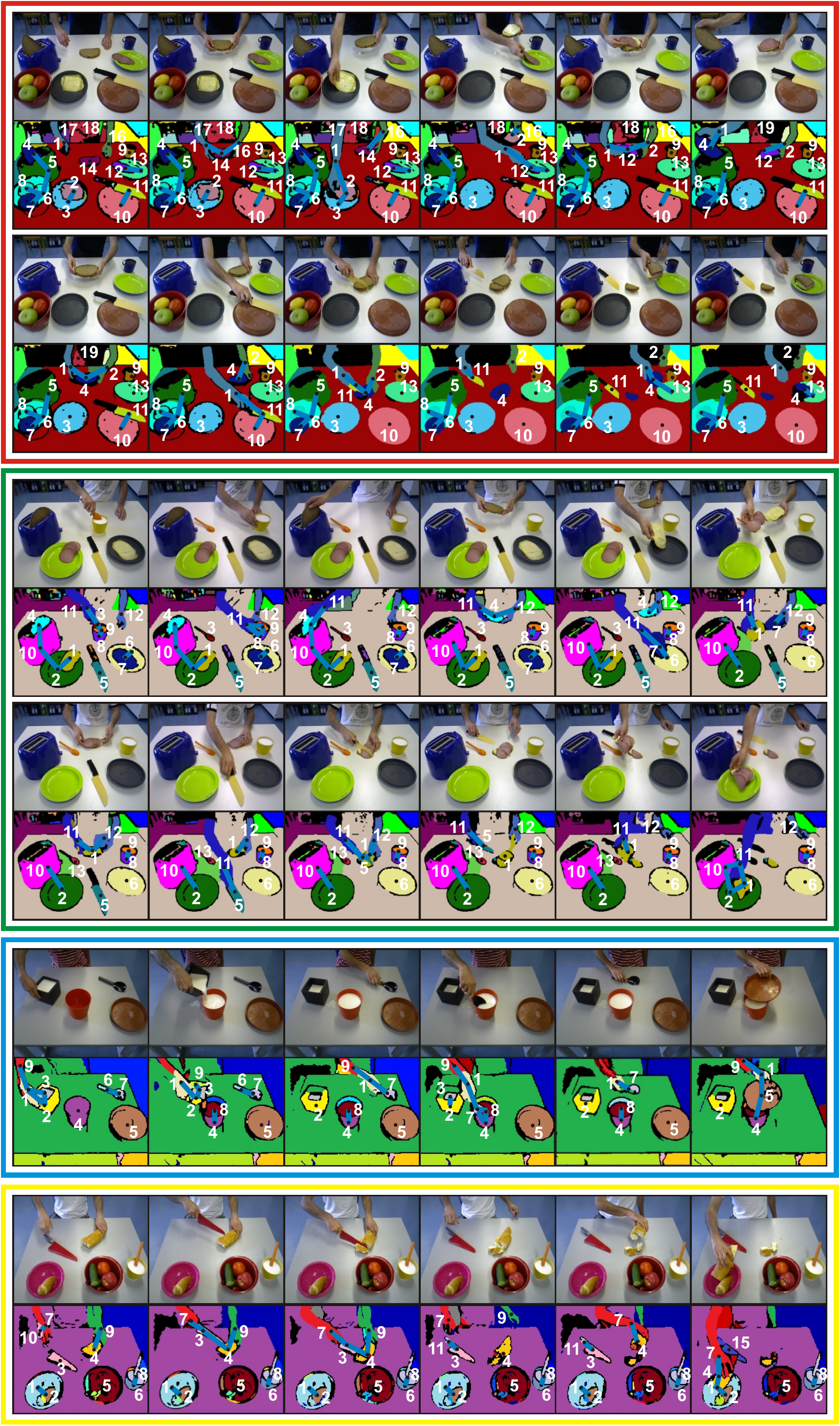}
  \caption{Sample frames with respective image segments and scene graphs from four different long chained manipulation sequences in the ManiAc dataset. In the given red, green, blue, and yellow frames subjects are performing different tasks such as ``{\it making a sandwich}", ``{\it preparing a breakfast}", ``{\it  pouring and stirring milk}", and ``{\it cutting and moving a piece of bread}". }
\label{fig:testchainedactions}
\end{figure}

The ManiAc dataset additionally provides $20$ long and complex chained manipulation sequences, such as ``{\it making a sandwich}", ``{\it preparing a breakfast}", ``{\it pouring and stirring milk}", or ``{\it cutting and moving a piece of bread}". These chained sequences have in total $103$ different versions of the learned $8$ single atomic manipulations as well as some novel tasks, such as {\it Pouring}. All these chained manipulations were performed in different orders, either sequentially or in parallel, with novel objects in various scene contexts to make the temporal segmentation and recognition steps more challenging.
Fig.~\ref{fig:testchainedactions} shows sample frames with tracked objects and scene graphs from four different chained manipulations to give an impression of the differences in the demonstrated long scenarios. This figure shows that scenes can be cluttered and graphs can also include more nodes some of which are occasionally changing labels due to occlusion problems. Even in such problematic cases, by applying the proposed top-down semantic reasoning we can not only decompose and recognize performed parallel or serial manipulation streams, but also extract manipulated objects by solely considering the semantics {\it (essence)} of manipulations.

All manipulations shown in this dataset were recorded with a single Microsoft Kinect sensor which provides both color and depth image sequences. Note that colored objects are preferred to cope with  the intrinsic limitations of the Kinect device.

\begin{figure*}[!t]
\centering
 \includegraphics[scale=0.6]{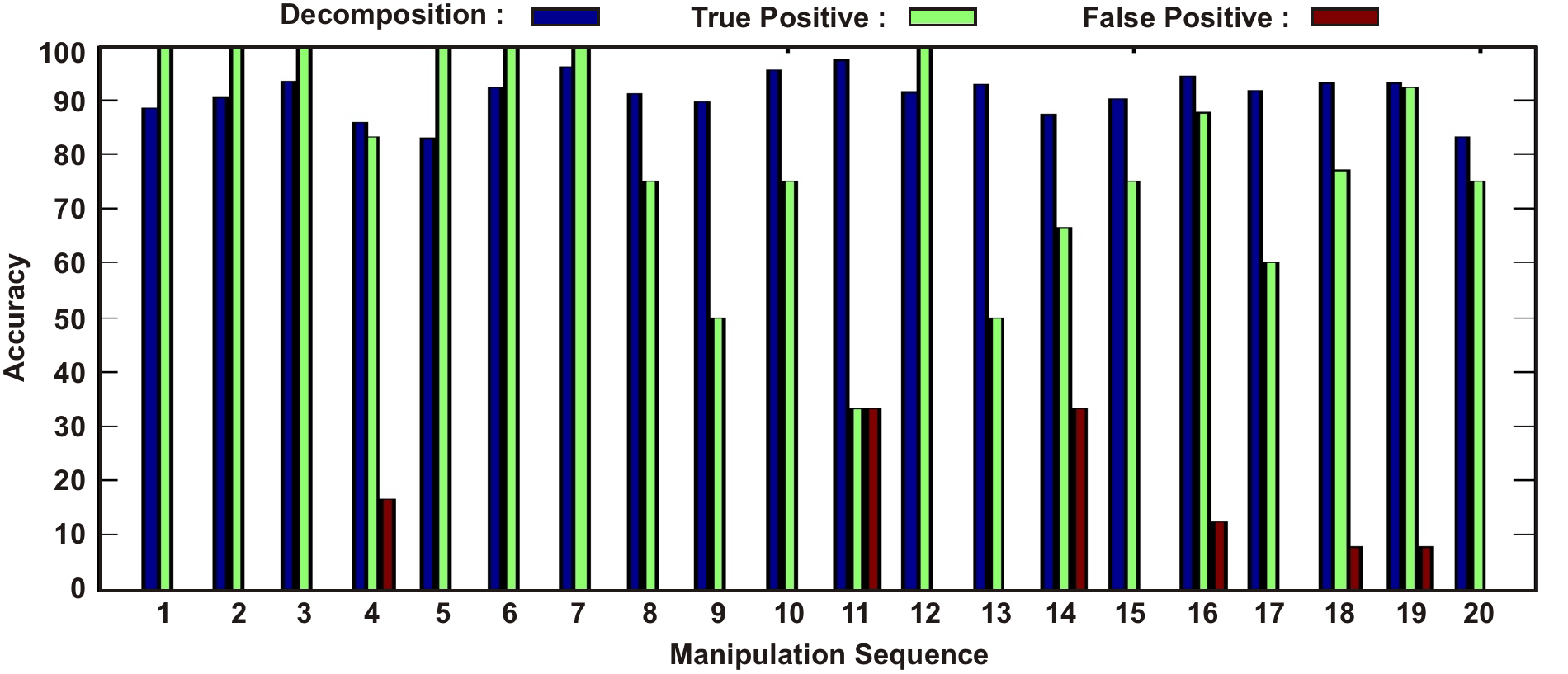}
  \caption{Manipulation temporal segmentation accuracy together with the true and false positive recognition rates of the detected manipulations in  $20$ chained long sequences in the ManiAc dataset.}
\label{fig:decompositionaccuracy}
\end{figure*}

In the first stage, we analyzed the temporal segmentation accuracy of the $20$ chained manipulation sequences in the ManiAc dataset. As described in section~\ref{sec:manipulationdecomposition}, the temporal segmentation process is bootstrapped with the estimation of the main {\it manipulator} in the scenario. We acquired $100\%$ correct {\it manipulator} estimation rate in all chained sequences. This obtained highest precision also leads to  robust semantic manipulation temporal segmentation. Frame-wise temporal segmentation accuracy was next computed by comparing our results with the human defined ground truth. The blue bars in Fig.~\ref{fig:decompositionaccuracy} indicate the final temporal segmentation rates successfully computed from each chained sequence. We obtained $91\%$ mean temporal segmentation accuracy over all  $20$ sequences. The reason of the little deviation from the ground truth is that the noisy segmentation and tracking information delay the detection time of the spatial relations (\ie touching events) between the {\it manipulator} and the {\it primary objects} in the scene. Such delays, however, do not corrupt the recognition phase as described below.

\begin{figure*}[!b]
\centering
 \includegraphics[scale=0.75]{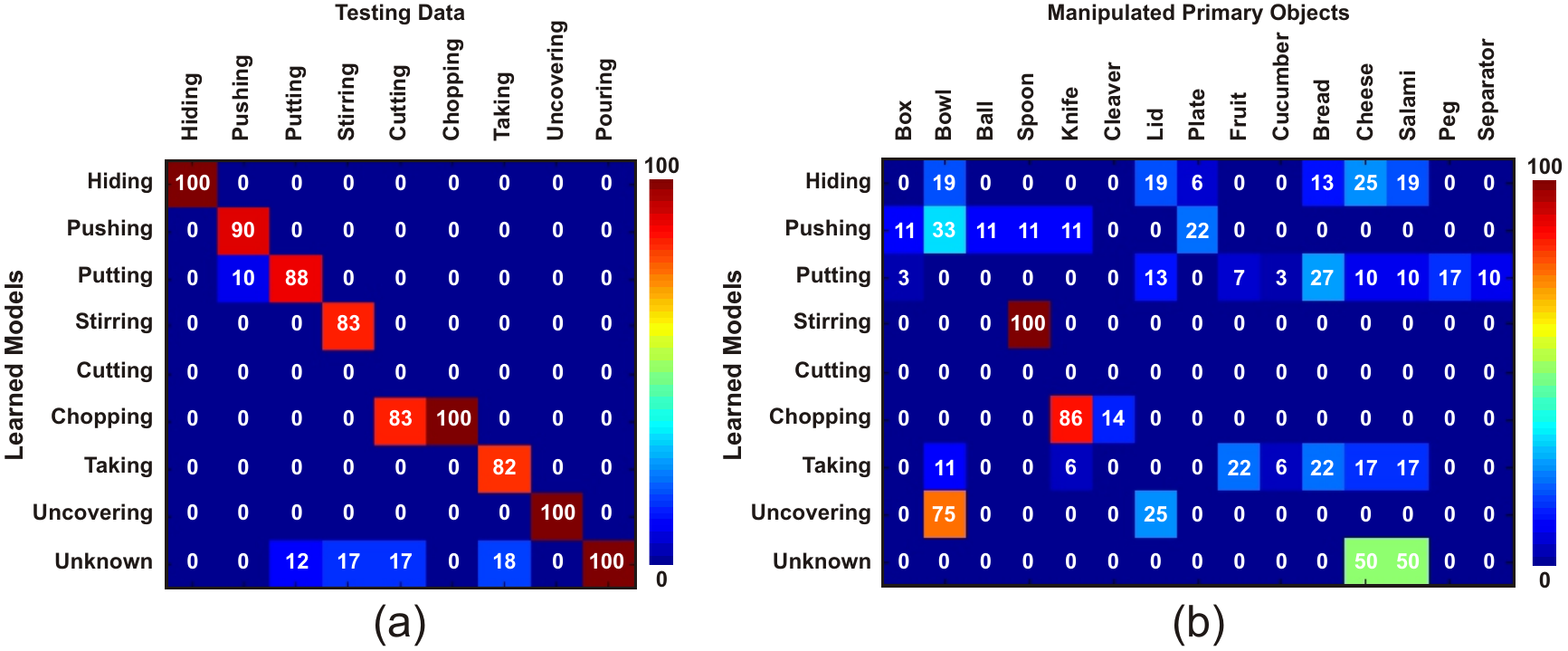}
  \caption{Confusion matrix showing (a) the manipulation recognition accuracies of the tested manipulation types embedded in the $20$ chained sequences and (b) the usage rate of different {\it primary objects}  manipulated in the known manipulation types.}
\label{fig:recognitionaccuracy}
\end{figure*}

After the temporal segmentation process, we evaluated the recognition rates of the sequential and parallel manipulation streams existing in each chained sequence. As pointed out in section~\ref{sec:manipulationrecognition}, the recognition process is essentially based on the prediction of the {\it primary} and {\it secondary objects} and is followed with the comparison of the decomposed manipulation streams with the learned SEC models stored in the library. The green and red bars in Fig.~\ref{fig:decompositionaccuracy} depict the true and false positive recognition rates of the detected manipulation streams in each chained sequence when compared with the learned $8$ SEC models. We computed the mean true and false positive rates as $80\%$ and $6\%$, respectively.
There are two reasonable explanations for observing a slight drop in true positives and having a relatively small false positive rate in some sequences. First, the novel manipulation types (\eg~{\it Pouring}) in
the chained sequences have not previously been learned as SEC models.
The proposed framework treats such novel manipulations as {\it Unknown} if their semantic similarities with the already known models are less than the learned threshold $\tau_{sem}$ (introduced as $72\%$ in section~\ref{sec:manipulationrecognition}). Therefore, the framework exhibits a lower than $100\%$ true positive rate. The second, and the most important, factor is that the learned {\it Cutting} and {\it Chopping} models are semantically similar and thus are merged in the recognition phase. This is because both manipulations have the same fundamental action primitives, \ie similar columns in the event chains, and the only differences are mostly in the followed trajectories and velocity of the movements which are not captured by SECs. Hence, the framework naturally merges those two manipulation types, which leads to an increment in the false positive rates. This result is fully compatible with our previous findings shown in \cite{AksoyRAS2015}, in which SEC models are on-line learned without using any human intervention. The on-line learning framework in \cite{AksoyRAS2015} retrieved one single SEC model by naturally merging both {\it Cutting} and {\it Chopping} manipulation samples due to exploring high semantic similarity between each type. Note, there is no ``ultimate truth" with respect to what one might call semantically similar (or dissimilar) actions. We will discuss these conceptional problems at great length in the Discussion section. Here, we note that all our recognition results are intrinsically consistent and robust.

\begin{figure*}[!t]
\centering
 \includegraphics[scale=0.72]{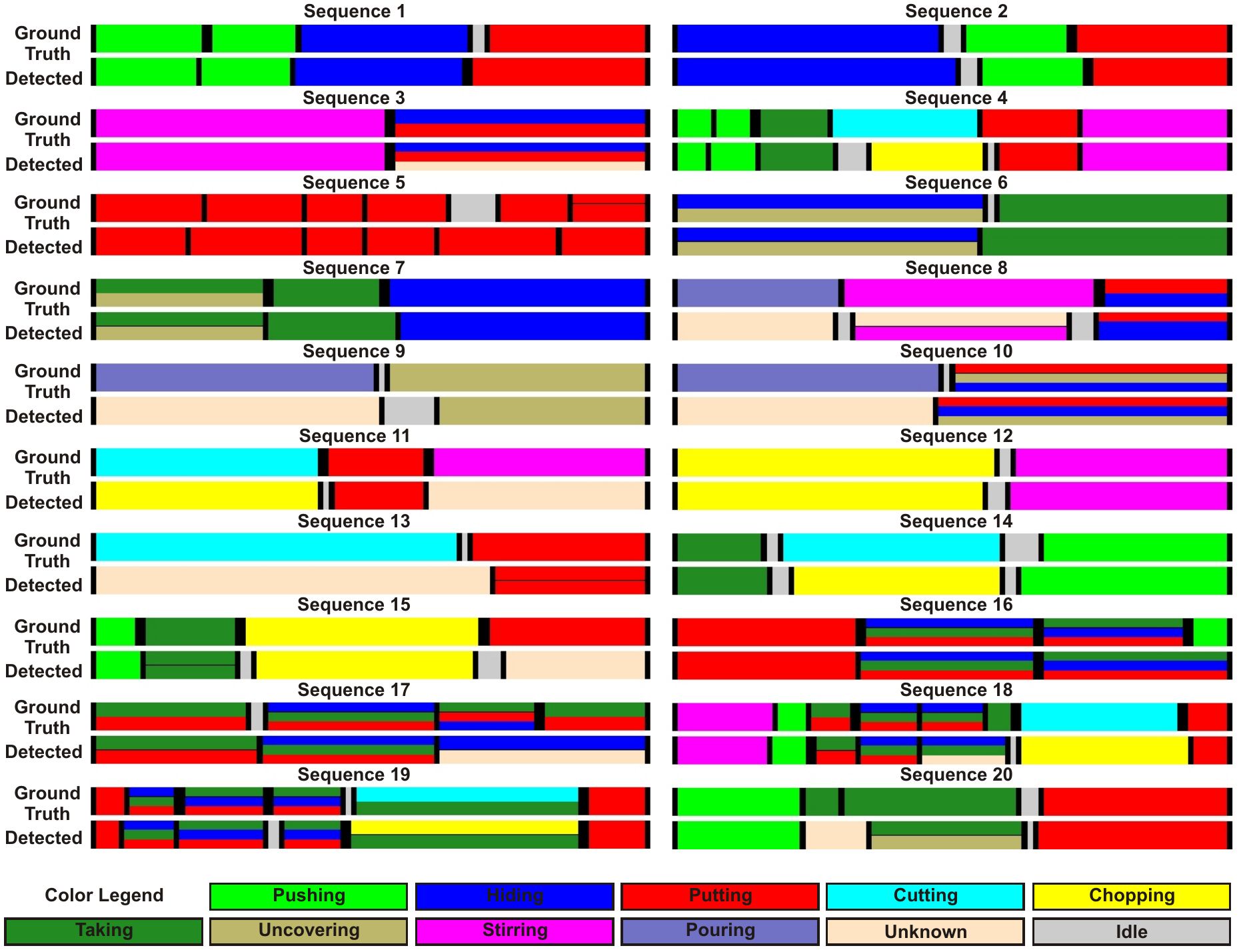}
  \caption{Automatic temporal segmentation and recognition results of the $20$ chained manipulation sequences versus human labeled ground truth. The action segments are color coded. Black frames indicate the border of each manipulation stream. Gray color represents the {\it Idle} actions in which the {\it manipulator} is not interacting with any object while switching from one manipulation to the next.}
\label{fig:decompositionvsgtruth}
\end{figure*}

To quantitatively evaluate this, Fig.~\ref{fig:recognitionaccuracy}~(a) shows a confusion matrix depicting the recognition accuracies of the $103$ tested manipulation samples, existing in the $20$ chained sequences, with respect to the learned $8$ SEC models. The first impression that the figure conveys is that the tested {\it Cutting} samples are naturally interpreted as   {\it Chopping}  because of the reason discussed above. However, no other confusion occurs, except an observed $10\%$  misclassification rate for the {\it Pushing} action.
It is also interesting to note that the novel {\it Pouring} manipulations demonstrated in the chained sequences were never confused with any of the known SEC models because of having a distinct semantics and, thus, were always classified as {\it Unknown}.

Fig.~\ref{fig:decompositionvsgtruth} displays the final temporal segmentation and recognition results of the chained sequences together with the human labeled ground truth. This side-by-side comparison shows that the system had successfully handle also the complicated parallel manipulation streams. Note that the lengths of chained sequences are normalized for the sake of clarity in the display.

 
\begin{table*}[!b]
\centering
\caption{Action detection rates with and without considering the role of \textit{secondary objects}. The first row identifies $20$ chained actions in the ManiAc dataset. The second and third rows respectively indicate the total number of single and parallel atomic actions in each chained action. $TP^{+}$ and $FP^{+}$ represent \textit{True Positive} and \textit{False Positive} rates when  \textit{secondary objects} are considered, whereas $TP^{-}$ and $FP^{-}$ show the results in the case of ignoring   \textit{secondary objects}. }
\begin{center}
\scalebox{0.6}{
\begin{tabular}{ |p{1.3cm}||p{0.7cm}|p{0.7cm}|p{0.7cm}| p{0.7cm}|p{0.7cm}|p{0.7cm}|p{0.7cm}|p{0.7cm}|p{0.7cm}| p{0.7cm}|p{0.7cm}|p{0.7cm}| p{0.7cm}|p{0.7cm}|p{0.7cm}|p{0.7cm}|p{0.7cm}|p{0.7cm}| p{0.7cm}|p{0.7cm}||p{0.7cm}|p{10.7m}|}
 \hline
                 &\centering 1  &\centering 2  &\centering 3  &\centering 4  &\centering  5 &\centering 6  &\centering 7  &\centering 8  &\centering 9  
&\centering 10   &\centering 11 &\centering 12 &\centering 13 &\centering 14 &\centering 15 &\centering 16 &\centering 17 &\centering 18 &\centering 19	
&\centering 20   &$\centering$Total\\ 
 
\hline
\centering \textit{Single} &\centering 4   &\centering 3  &\centering 1  &\centering 6 &\centering  5  &\centering 1  &\centering 2  &\centering 2  &\centering 2  
&\centering 1  &\centering 3 &\centering 2  &\centering 2  &\centering 3 &\centering  4  &\centering 2  &\centering 0  &\centering 5  &\centering 2	
&\centering 4  &$\centering$ 54\\ 

\centering \textit{Parallel} &\centering 0   &\centering 0  &\centering 2  &\centering 0 &\centering  3  &\centering 2  &\centering 2 &\centering 2 &\centering 0  
&\centering 3    &\centering 0   &\centering 0  &\centering 0  &\centering 0 &\centering  0  &\centering 6  &\centering 10 &\centering 8  &\centering 11	
&\centering 0    &$\centering$ 49\\ 
 
\hline

\centering $TP^{+}$ &\centering $100\%$  &\centering $100\%$ &\centering $100\%$ &\centering $83\%$  &\centering $100\%$ &\centering $100\%$ &\centering $100\%$ &\centering $75\%$  &\centering $50\%$   &\centering $75\%$  &\centering $33\%$  &\centering $100\%$  &\centering $50\%$  &\centering $67\%$  &\centering $75\%$  &\centering $87\%$  &\centering $60\%$   &\centering $77\%$  &\centering $92\%$  &\centering $75\%$   &$\textbf{80}\%$\\ 

\centering $TP^{-}$ &\centering $100\%$  &\centering $100\%$ &\centering $33\%$  &\centering $83\%$  &\centering $62\%$  &\centering $33\%$  &\centering $50\%$
&\centering $0\%$  &\centering $50\%$  &\centering $0\%$  &\centering $33\%$  &\centering $100\%$  &\centering $0\%$  &\centering $67\%$  &\centering $50\%$  &\centering $12\%$  &\centering $0\%$    &\centering $23\%$  &\centering $15\%$  &\centering $50\%$   &$43\%$\\

\hline  
   
\centering $FP^{+}$ &\centering $0\%$    &\centering $0\%$   &\centering $0\%$   &\centering $17\%$  &\centering $0\%$   &\centering $0\%$   &\centering $0\%$ &\centering $0\%$   &\centering $0\%$   &\centering $0\%$  &\centering $33\%$  &\centering $0\%$   &\centering $0\%$  &\centering $33\%$  &\centering $0\%$  &\centering $12\%$  &\centering $0\%$   &\centering $8\%$  &\centering $8\%$   &\centering $0\%$   & $\textbf{5.5}\%$\\ 

\centering $FP^{-}$ &\centering $0\%$    &\centering $0\%$   &\centering $0\%$   &\centering $17\%$  &\centering $0\%$   &\centering $0\%$   &\centering $0\%$ &\centering $0\%$   &\centering $0\%$   &\centering $0\%$  &\centering $33\%$  &\centering $0\%$   &\centering $0\%$  &\centering $33\%$  &\centering $0\%$  &\centering $0\%$   &\centering $0\%$   &\centering $8\%$  &\centering $0\%$   &\centering $0\%$   &$4.5\%$\\ 
     
\hline
\end{tabular}
}
\end{center}
\label{tab:secobjcontribution}
\end{table*}

As also addressed in section~\ref{sec:manipulationrecognition}, without requiring any object recognition framework we can correctly derive the {\it primary} and {\it secondary objects} utilized in the perceived manipulations. Fig.~\ref{fig:recognitionaccuracy}~(b) indicates the estimated {\it primary object} types that were frequently manipulated in the detected manipulation samples from the $20$ chained sequences.
For instance, {\it Spoon} was the only object type primarily employed in the detected {\it Stirring} manipulations, whereas {\it Knife} and {\it Cleaver} were heavily preferred in the {\it Cutting} and {\it Chopping} tasks. On the other hand, other object kinds like  {\it Bread}, {\it Cheese}, and {\it Salami} were utilized in the {\it Taking}, {\it Putting}, and  {\it Hiding} samples. This is because the scenarios such as ``{\it making a sandwich}" or ``{\it preparing a breakfast}" required taking  cheese or bread slices and putting them on top of each other, which naturally resulted in the disappearance of some objects correctly interpreted as hiding. These findings verify that the proposed framework can also automatically discover the link between actions and objects. Results on the likewise estimated {\it secondary objects} are not shown here to save some space. Taken together this clearly demonstrates (Fig.~\ref{fig:recognitionaccuracy}) a very high recognition rate for actions and objects in the long and complex ManiAc dataset and the next section shows that this also holds for other data.

\begin{figure*}[!t]
\centering
 \includegraphics[scale=0.6]{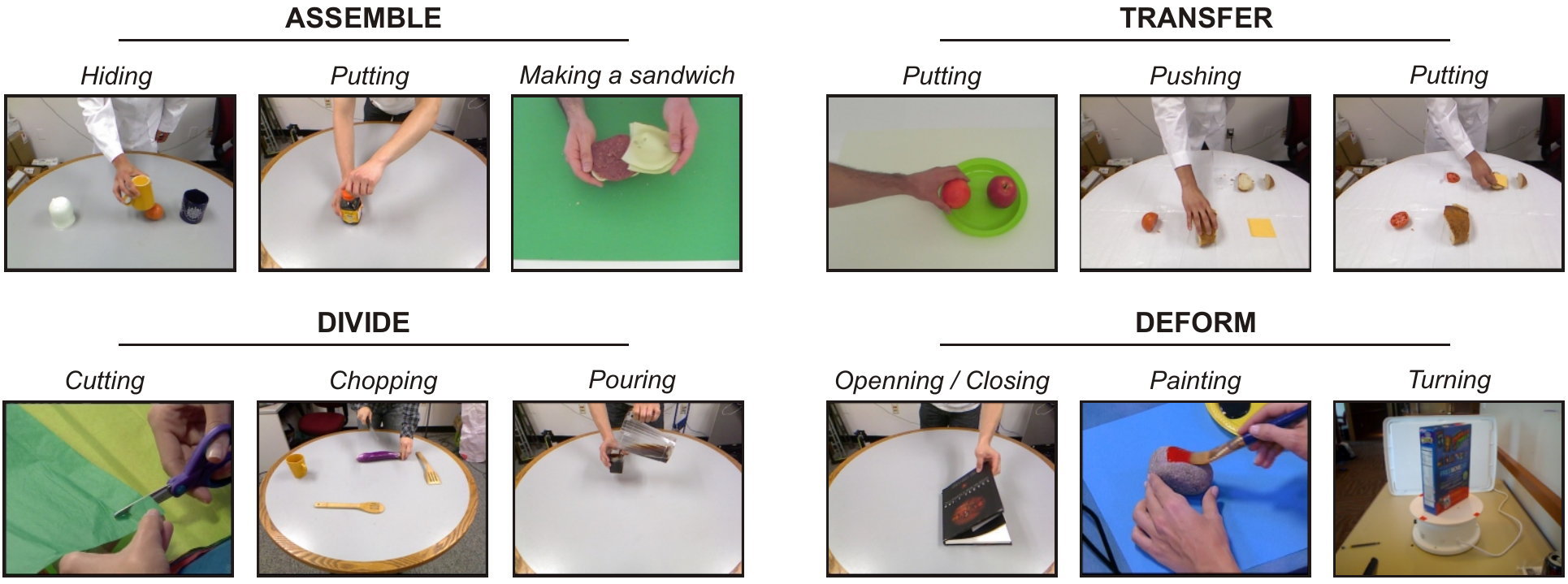}
  \caption{Sample frames from the four different manipulation categories in the MAC dataset.}
\label{fig:mac_actions}
\end{figure*}

\begin{figure*}[!b]
\centering
 \includegraphics[scale=0.78]{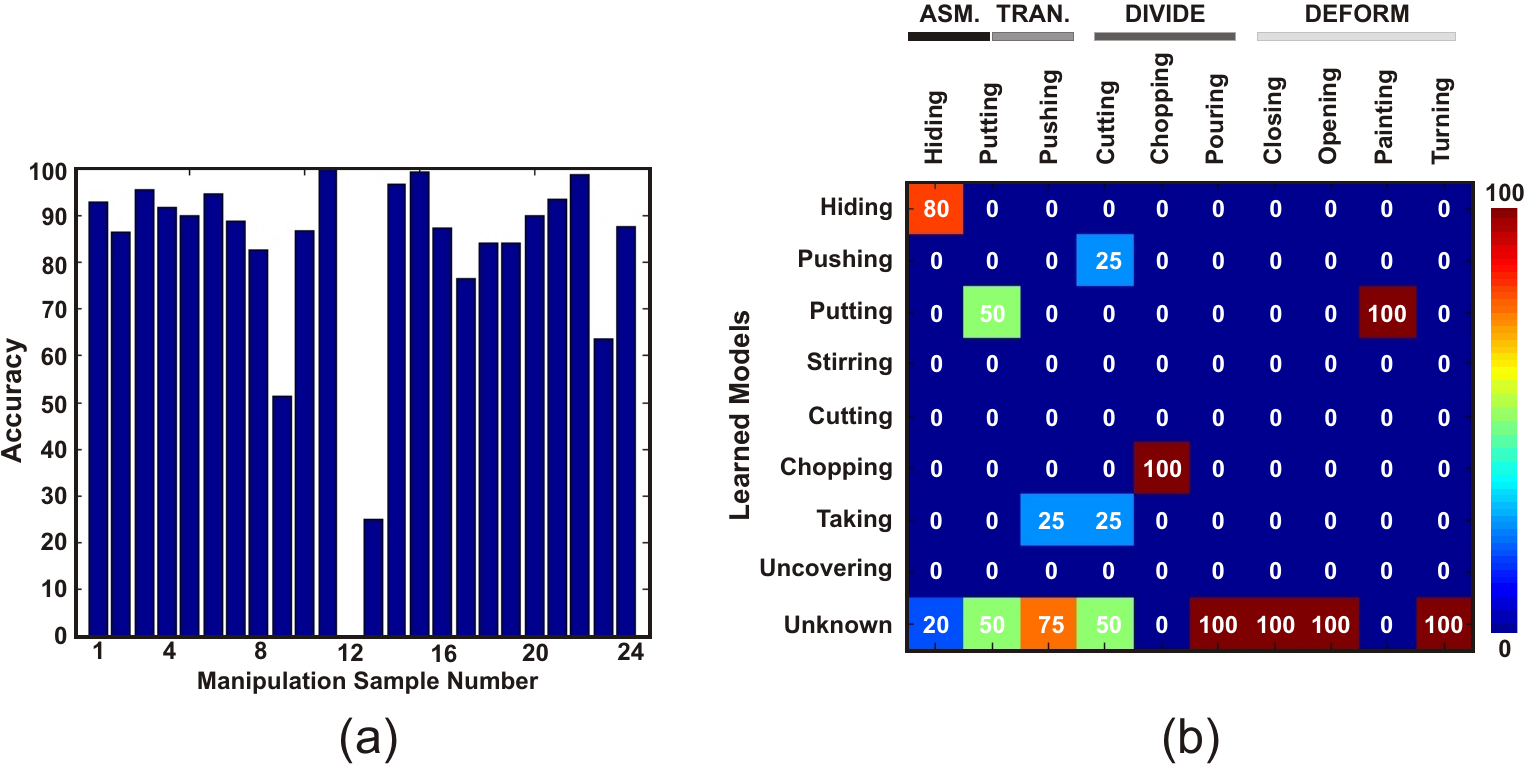}
  \caption{Experimental results from the MAC dataset. (a) Manipulation temporal segmentation accuracies. (b) Recognition accuracies of the decomposed manipulations with respect to the same $8$ SEC models learned from the ManiAc dataset. }
\label{fig:mac_results}
\end{figure*}

In order to address the main contribution of the \textit{secondary object} estimation, we repeat the detection of parallel actions without taking the role of \textit{secondary objects} into account. Table~\ref{tab:secobjcontribution} shows the total number of single and parallel actions embedded in $20$ chained actions together with the average true and false positive rates in the cases of including and excluding the role of \textit{secondary objects}.
We observe $80\%$ true positive rate ($TP^{+}$) once   \textit{secondary objects} are estimated as proposed in section~\ref{sec:manipulationrecognition}.
On the other hand, when the exploration of \textit{secondary objects} is omitted, the overall average accuracy (mean of true positives, \ie $TP^{-}$) drops to $43\%$  as depicted in the very last column in Table~\ref{tab:secobjcontribution}. The main drop in the accuracy particularly occurs in cases of having parallel action streams. For instance, activity numbers $16$ and $17$ in Table~\ref{tab:secobjcontribution} have more parallel actions involved, hence, correctly detected action rates dramatically drop. If the manipulation activity is composed of only single atomic actions, \eg first and second chained actions, the average accuracy remains the same.  This results support the claim that \textit{secondary objects} play a crucial role only in detecting parallel actions.
It is also important to note that the rate of false positives ($FP^{-}$) slightly decreases to $4.5\%$ from $5.5\%$. 
This little drop is a very important finding indicates that parallel actions are treated as {\it Unknown} rather than being misclassified once \textit{secondary object} are neglected. This also reveals the robustness of our action recognition method.

\subsection{Manipulation Action Consequences (MAC) Dataset}
\label{sec:macdataset}

The recently published MAC dataset \citep{Yang13} contains in total $24$ manipulation demonstrations categorized under four different action types: {\it ASSEMBLE},  {\it TRANSFER},  {\it DIVIDE}, and  {\it DEFORM}. Each category has $6$ various samples which were recorded with either a single Kinect device or a regular  RGB camera. These $24$ demonstrations consist of a total of $31$ single atomic manipulations, some of which were presented as chained sequences with sequential and parallel manipulation streams (\eg~{\it Making a sandwich}). Fig.~\ref{fig:mac_actions} displays  sample images from various scenarios in each manipulation category to indicate the level of differences between the demonstrated tasks.

\begin{figure*}[!t]
\centering
 \includegraphics[scale=0.8]{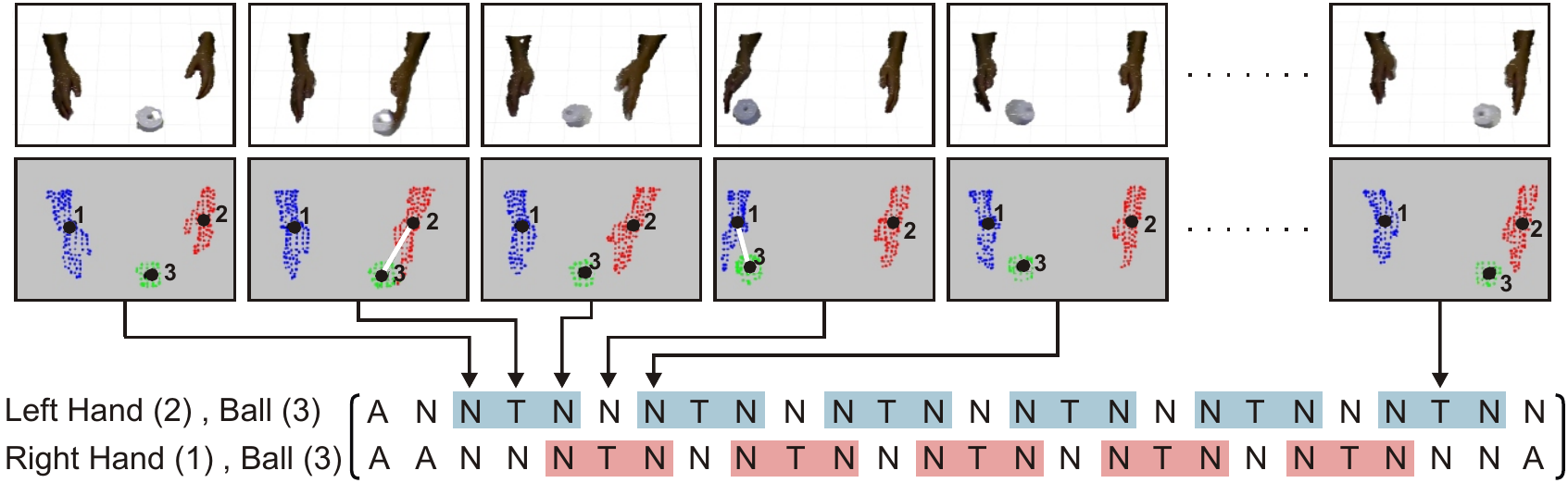}
  \caption{The event chain representation for the two-hand {\it Pushing} action sequence in the MOT dataset. The blue and red blocks in the SEC highlight decomposed {\it pushing} actions performed by the left and right hands, respectively. On the top sample original images with respective objects (colored regions) and main graphs are displayed.
 }
\label{fig:mot_pushing}
\end{figure*}

Since the the MAC dataset is very problematic due to missing depth information and huge changes between hand movements in consecutive frames (\ie frame dropping problem), we bypassed the segmentation and tracking phases and created event chains in a supervised manner using human expertise. As one of the biggest advantages of our proposed semantic segmentation and recognition framework, the cognitive systems here do not additionally require a new exhausting training period to test this novel dataset since the concept of semantics yields always the same essence of the manipulations, which does not dramatically alter with new manipulation observations. This allows us to employ the same SEC models, learned from the ManiAc dataset, in order to recognize the decomposed manipulations in the MAC dataset.

Fig.~\ref{fig:mac_results}~(a) depicts the temporal segmentation results of the $24$ manipulations in the MAC dataset. The mean temporal segmentation accuracy was computed as $81\%$ over all  $24$ sequences. Because some manipulation samples in the MAC dataset do not adhere to the action descriptive rules introduced in section~\ref{sec:manipulationdecomposition}, the temporal segmentation accuracy  dramatically dropped, for example, for manipulation sample number $12$.
This action consists of a box rotating on a turntable as depicted by the last frame in Fig.~\ref{fig:mac_actions}. This is not a manipulation in any  sense and, thus, this action is outside the scope of this paper. Therefore, our proposed framework treats these types of actions as {\it Unknown}. Fig.~\ref{fig:mac_results}~(b) displays the final recognition success of all decomposed $31$ atomic manipulations embedded in the $24$ demonstrations. The novel manipulation types, such as {\it Pouring}, {\it Opening}, and {\it Closing} were all correctly treated as {\it Unknown} due to having unique and distinct semantics compared to the previously learned 8 SEC models. On the other hand, the {\it Painting} sample, which is not existing in the learned SEC models, was interpreted as {\it Putting}. This is an entirely correct reasoning since the paint can be treated as an object that is being put on some other objects. (Again we point to the discussion section for an account on ``semantic similarities".) Although almost all manipulations were recognized with high accuracy, the {\it Pushing} and {\it Cutting} samples were  misinterpreted.

\subsection{Multiple Object Tracking (MOT) Dataset}
\label{sec:mot_dataset}

\begin{figure*}[!t]
\centering
 \includegraphics[scale=0.8]{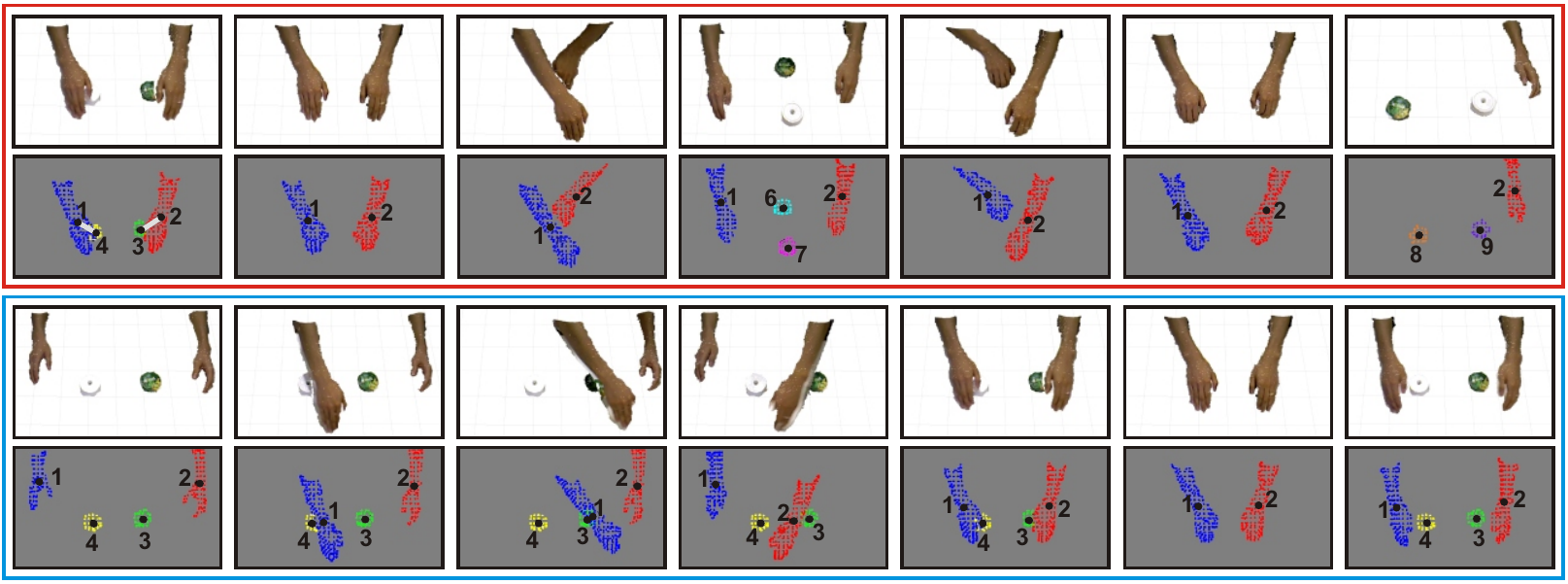}
  \caption{Two-hand {\it Occluding} actions from the MOT dataset. In the red frame, object labels of the occluded and replaced objects are altering, which leads to failures in the temporal segmentation phase. In the blue frame, both hands are occluding objects without carrying out any specific task, hence,  {\it occluding} actions were recognized as {\it Unknown}.}
\label{fig:mot_occlusion}
\end{figure*}

The MOT dataset  was recently introduced in \cite{Koo14} to investigate the multiple object tracking problem by applying dynamically updated object models without employing any prior knowledge. The dataset consists of $8$ different scenarios with a total of $23$ atomic actions. The scenarios demonstrate both single- and two-hand chained manipulation sequences, such as {\it Pushing}, {\it Stacking}, {\it Unstacking}, and {\it Occluding}. Fig.~\ref{fig:mot_pushing} and~\ref{fig:mot_occlusion} display sample frames from the {\it Pushing} and {\it Occluding} scenarios.

Since the MOC dataset already provides the segmented  scene configuration in $3D$, we bypassed our image segmentation and tracking step (section~\ref{sec:segmentationandtracking}) and started directly with the event chain extraction. To cope with multiple hands in the manipulation, we biased the {\it manipulator} estimation step (section~\ref{sec:extractingmanipulator}) with the actual number of hands in the scene. Given  $K$ as number of the {\it manipulators}, we computed the probability values from Eq.~{\ref{prob_segments}} for all possible combinations of objects, \ie graph nodes in the SEC, taken $K$ at a time. The object combination with the  highest probability value, \ie longest $[N, T, \cdots, T, N]$ sequence, was then considered as representing the {\it manipulator}. Fig.~\ref{fig:mot_pushing} illustrates the two-hand {\it Pushing} action sequence from the MOT dataset together with the extracted SEC representation and some sample key frames. Object numbers $1$ and $2$  were correctly estimated as the main {\it manipulators} and the extracted event chain was accordingly broken up into pieces, each of which is highlighted with the red and blue blocks in Fig.~\ref{fig:mot_pushing}.

\begin{figure}[!b]
\centering
 \includegraphics[scale=0.67]{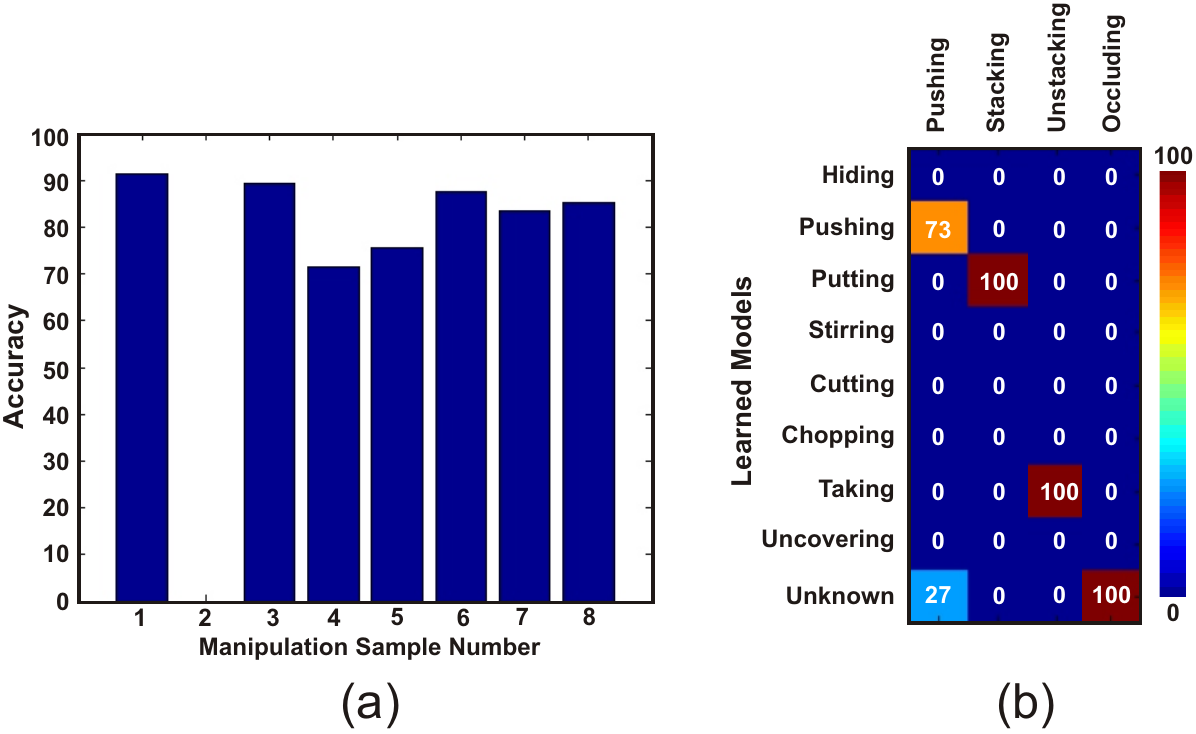}
  \caption{Experimental results from the MOT dataset. (a) Manipulation temporal segmentation accuracies. (b) Recognition accuracies of the decomposed manipulations with respect to the same $8$ SEC models learned from the ManiAc dataset.  }
\label{fig:mot_results}
\end{figure}

Fig.~\ref{fig:mot_results}~(a) depicts the final temporal segmentation accuracies for each of the  $8$ sequences in the MOT dataset. The drop in the second manipulation sample is due to the inconsistencies in the tracking phase, \ie due to a {\it segment discontinuity} problem. In this second scenario, for instance, two hands were moving objects around after occluding them completely. At each time the hands withdrew, the replaced objects emerged with novel object labels as illustrated in the red frame in Fig.~\ref{fig:mot_occlusion}. As the object labels altered unpredictably, the event chain representation could not capture any sequence of $[N, T, \cdots, T, N]$ relations, required for the manipulator estimation as described in section~\ref{sec:extractingmanipulator}, which led to failures in the temporal segmentation phase. If we exclude this second manipulation sample, the mean temporal segmentation accuracy reaches $83\%$. This result underlines the fact that our proposed semantic segmentation method is also suitable for manipulations with multiple {\it manipulators} as long as the scene objects are consistently trackable.

Fig.~\ref{fig:mot_results}~(b) depicts the recognition successes of all $23$ atomic actions performed either sequentially or in parallel in the $8$ different scenarios. As in the case of the MAC dataset, we here again employed the SEC models, learned from the ManiAc dataset, during the process of manipulation recognition. In the monitoring stage, some versions of the {\it Pushing} manipulations were missed out because of the same {\it segment discontinuity} problem pointed out above. On the other hand, all versions of the {\it Occluding} demonstrations were correctly recognized as {\it Unknown} since in such actions the hands were indeed not applying any certain task on the objects, but were rather performing random movements (to verify the stability in the segmentation). Thus, hands and objects were here not interacting to issue any {\it touching} event required by the action descriptive rules summarized by Eq.~\eqref{ntn_sequence}. The blue frame in Fig.~\ref{fig:mot_occlusion} depicts sample frames from a version of the {\it Occluding} scenario, in which two hands are randomly moving above the objects without aiming at any specific task. Results derived from the MOT dataset consequently confirm the scalability of our proposed temporal segmentation and recognition framework to multi-hand manipulation actions even with the flexibility of replacing some submodules, such as the segmentation and tracking method.

\subsection{Baseline Experiments}
\label{sec:baselineexperiments}

In our baseline experiments, we used appearance and trajectory based state-of-the-art action descriptors, such as Space-Time Interest Points (STIP), Dense Trajectories (DT) and Improved Dense Trajectories (IDT). 
STIPs described in \citep{STIP2005} are local image points around which the image values exhibit significant structural variations in both space and time domains. 
%
%
DT was introduced in \citep{Wang2011} and computes various dense features such as static apprearance information, local motion information and relative motion between pixels. 
IDT  \citep{Wang2013} is the extended version  of DT by taking into account camera motion to remove false trajectories consistent with camera motion.

\begin{figure}[!b]
\centering
 \includegraphics[scale=0.6]{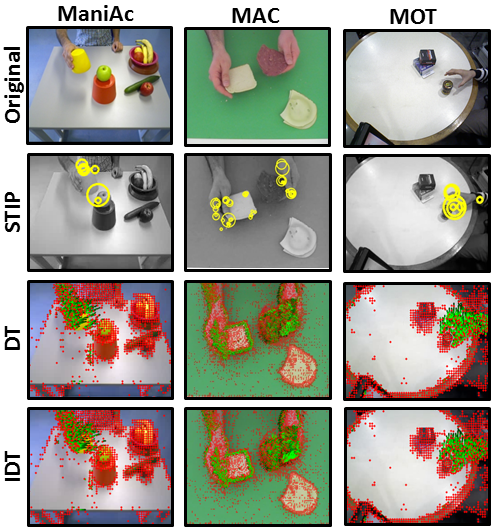}
  \caption{Sample images with detected STIP, DT and IDT features. }
\label{fig:baseline_features}
\end{figure}

In all our baseline experiments with STIP, DT and IDT we used the default parameters coming with the publicly available source codes. We computed Fisher vectors together with Gaussian Mixture Models in order to create the visual vocabulary from STIP features. In the case of DT and IDT descriptors,  we implemented a standard bag-of-features representation to construct a codebook. We clustered extracted DT and IDT features using K-means into a codebook of $400$ words, which is the same size for the STIP feature vocabulary. Detected DT and IDT descriptors were then assigned to their closest word in the codebook by considering the Euclidean distance. All detected STIP, DT, and IDT action descriptors were then passed to Support Vector Machines (SVM) in a one-versus-all fashion. It is here important to note that we performed several tests on various sets of codebook sizes with different SVM kernels (\eg linear and Chi2) and reported the highest results out of these experimental evaluations. 
Fig.~\ref{fig:baseline_features} illustrates detected STIP, DT, and IDT features on some samples frames from different datasets.

\begin{table*}[!b]
\centering
\caption{Classification performance comparison among different methods on three various datasets. Pr, Rc, and FS stand for Precision, Recall, and F-Score, respectively.}
\begin{center}
\scalebox{0.75}{
\begin{tabular}{ |p{5.5cm}||p{1.4cm}|p{1.4cm}|p{1.4cm}|| p{1.4cm}|p{1.4cm}|p{1.4cm}||p{1.4cm}|p{1.4cm}|p{1.4cm}| }
 \hline
 	   	       & \multicolumn{3}{|c||}{ManiAc \citep{AksoyRAS2015}} & \multicolumn{3}{|c||}{MAC \citep{Yang13}}  & \multicolumn{3}{|c|}{MOT \citep{Koo14}} \\
 \hline
               &	 \centering Pr&	\centering Rc&	\centering	FS & \centering	Pr & \centering Rc & \centering FS & \centering Pr & \centering Rc & $\centering$ FS  	\\ 
 \hline
 STIP \citep{STIP2005}  &\centering $20.6\%$ &\centering $31.8\%$ &\centering $25.0\%$ &\centering $36.6\%$ &\centering $24.0\%$ &\centering $29.0\%$ 
 &\centering $46.6\%$   &\centering $31.6\%$ &  $37.7\%$ \\
 
 DT   \citep{Wang2011}  &\centering $10.1\%$ &\centering $4.7\%$ &\centering $6.4	\%$ &\centering $12.5\%$ &\centering $20.8\%$ &\centering $15.6\%$ 
 &\centering $59.1\%$   &\centering $57.8\%$ &  $58.4\%$ \\
 
 IDT   \citep{Wang2013} &\centering $13.7\%$ &\centering $14.0\%$ &\centering $13.9\%$ &\centering $12.5\%$ &\centering $20.0\%$ &\centering $15.3\%$ 
 &\centering $18.3\%$   &\centering $17.2\%$ &  $17.7\%$ \\
 
 SECs (\textit{Ours})   &\centering $\textbf{91.8\%}$ &\centering $\textbf{91.0\%}$ &\centering $\textbf{91.4\%}$ &\centering $\textbf{52.6\%}$ 
 &\centering	$\textbf{53.1\%}$ &\centering $\textbf{52.8\%}$ &\centering $\textbf{93.3\%}$ &\centering $\textbf{87.5\%}$ &  $\textbf{90.3\%}$ \\ 

 \hline
\end{tabular}
}
\end{center}
\label{tab:baselineresults}
\end{table*}
  
\begin{figure}[!t]
\centering
 \includegraphics[scale=0.2]{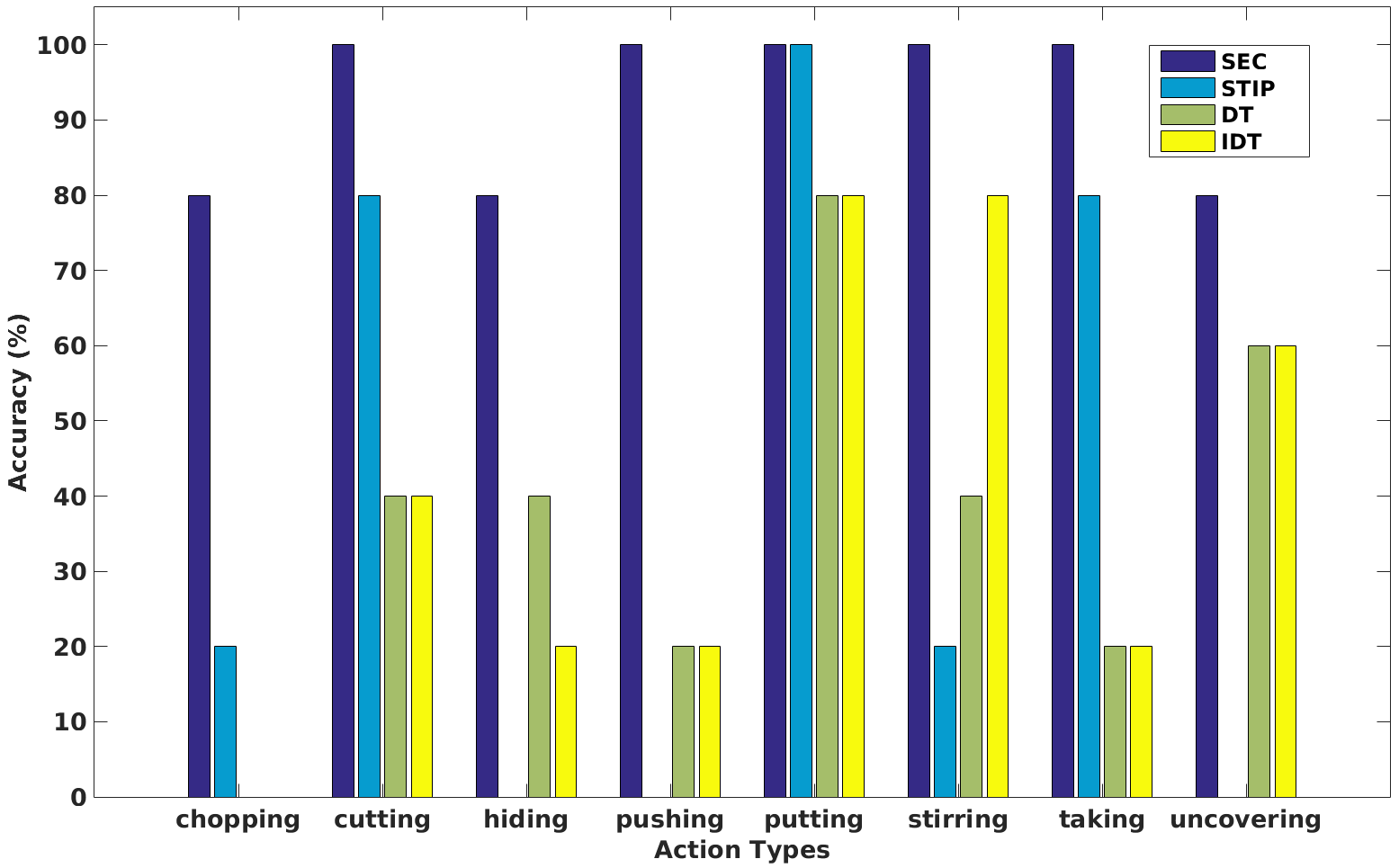}
  \caption{Classification accuracies of SEC, STIP, DT, and IDT based  methods on the ManiAc dataset. }
\label{fig:baseline_validation}
\end{figure}

In our first baseline experimental set up, we investigated the discriminative power of these local visual descriptors, \ie STIP, DT and IDT, and compared with our SEC based approach. For this task, we used $120$ demonstrations of $8$ different atomic action types provided in the ManiAc dataset (see Section~\ref{sec:maniacdataset}). Since each action class has $15$ different versions, we employed the first $10$ samples for  training  and the rest for testing. A separate SVM classifier was defined for each class type. 
Fig.~\ref{fig:baseline_validation} presents a per-class test score after performing the same training procedure for all four methods. Here, each bar in the plot shows the classification accuracy ($\%$) referring  to the average number of true positive predictions out of total tested sample number. 
This first experiment shows that local visual descriptors do not perform well and have poor performances on some specific action types, such as \textit{Chopping}, whereas our proposed SEC method has the highest score for all action types. 
Over eight classes we obtained $92.5\%$ average classification rate for the SEC method, whereas it was $37.5\%$, $37.5\%$, and $40\%$ for STIP, DT, and IDT methods, respectively.

Next, we would like to address the problem of transferring the learned codebooks, \ie visual vocabularies, across different datasets. In this task, we trained the same four classifiers, \ie SEC, STIP, DT, and IDT based approaches, with  all $15$ versions provided for each action type in the ManiAc dataset. In the testing phase, we first used $20$ long action sequences coming with the ManiAc dataset (see Section~\ref{sec:maniacdataset}). We then measured the performance of all these already trained classifiers on the MAC and MOT datasets described in  Section~\ref{sec:macdataset} and  Section~\ref{sec:mot_dataset}, respectively. 
We here note that these experiments essentially assess the action recognition power of the state-of-the-art methods which can not perform temporal action segmentation. Hence, we provided manually segmented actions for STIP, DT and IDT methods, whereas we let our SEC framework automatically segment all these long activities as explained in Section~\ref{sec:decomposingsecs}. In the computation of precision and recall values in all methods, we treated \textit{Cutting} and \textit{Chopping} actions as type-similar due to having high semantic structures. Different from conventional mehtods, in the SEC approach, we also introduced \textit{Unknown} class to assess zero-shot action recognition performance for novel action types.
 

Table~\ref{tab:baselineresults} shows the average classification performance comparison among the aforementioned methods on three datasets. We again obtained quite low performances with the STIP, DT and IDT based classifiers. Our proposed SEC method has the highest precision and recall values and outperforms these state of the art methods. This empirical result shows that conventional approaches are far away from the generalization perspective.

Both baseline experiments indicated in Fig.~\ref{fig:baseline_validation} and Table~\ref{tab:baselineresults} demonstrate that employing the semantic information yields more accurate recognition performance compared to what the conventional approaches can achieve.
This is likely because such conventional action descriptors highly depend on the scene context and followed trajectory patterns which can vastly alter from one demonstration to another as it is the case for the here benchmarked datasets. 
For instance, as depicted in Fig.~\ref{fig:training_cutting_versions} each of $8$ atomic actions in the ManiAc dataset was performed by five different persons each followed different trajectories by using several objects in different scene contexts. This indeed poses a challenge for the recognition task. 
Furthermore, in the long activities coming with all three dataset, individual manipulation streams were performed either sequentially or concurrently at varying speeds in more cluttered scenes (see Fig.~\ref{fig:testchainedactions}). Thus, optical flow features even for the same activity  have significant changes.
As also reported in several other papers \cite{Li_2015_CVPR,Fathi2011} local spatio-temporal features required by conventional methods are often captured at  locations irrelevant to a performed action due to wrongly computed optical flow information.
In this regard, it is more likely that the conventional models received vastly different and sparse optical flow signals from the trained and tested action sets.

Unlike our SEC approach, traditional action classifiers cannot detect overlapping actions. Those missed actions also lead to significant drops in the final accuracy computation.

Another reason of having the low performance with conventional methods is that the test data set involved novel action types, such as \textit{Pouring} and \textit{Painting}  (see Fig.~\ref{fig:testchainedactions} and Fig.~\ref{fig:mac_actions}) for which we lack pre-trained action models in order to test zero-shot action recognition performance. However, the main aim of these conventional action descriptors is to capture descriptive key features that are relevant to trained actions, which are in general short actions. Therefore, those unseen actions were misclassified by these methods.

\begin{figure*}[!b]
\centering
 \includegraphics[scale=0.6]{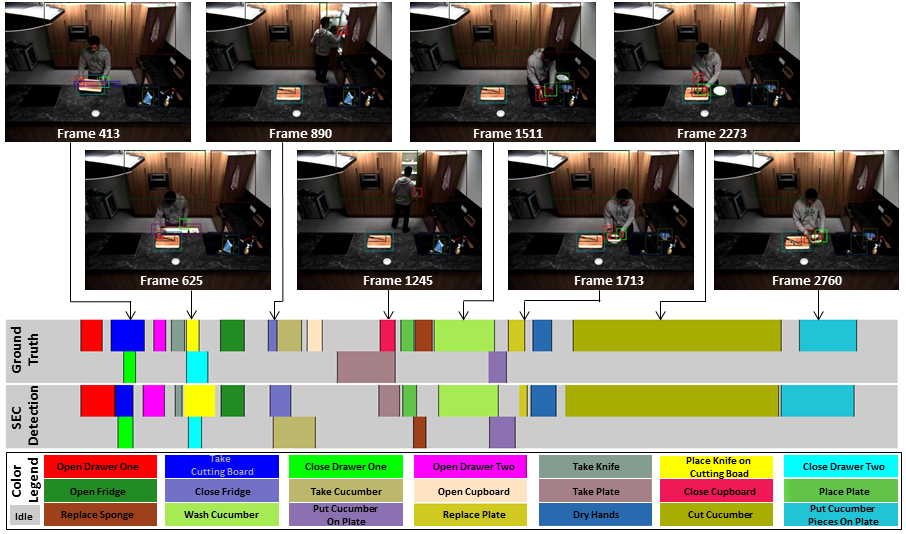}
  \caption{\red{Temporal segmentation result for a sample MPII cooking activity. The action segments are color coded. Bounding boxes on each frame represent the tracking results.}}
\label{fig:mpIIDecompImage}
\end{figure*}

Consequently, conventional feature  extractors are essentially based on appearance and motion in certain space time volumes (intervals). Most of those techniques computes local space-time gradients. As baseline experimental results highlight such approaches are not sufficient for modeling variations even in the same action type. 
Different from these  unified descriptors, the here presented SEC framework can handle variations in the performed actions and detect untrained action types as \textit{Unknown} as depicted in   Figs.~\ref{fig:recognitionaccuracy},~\ref{fig:mac_results}, and~\ref{fig:mot_results}.
The data presented in Fig.~\ref{fig:baseline_validation} and Table~\ref{tab:baselineresults} are clear indications of the scalability and stability features of our proposed SEC approach. Furthermore, Table~\ref{tab:baselineresults} clearly supports the generalization capacity and the transferability of the learned SEC representations. 
In contrast to the state-of-the-art action descriptors, the proposed SEC approach can also automatically execute temporal action segmentation  and  additionally categorize manipulated objects according to their performed roles, all in the same coherent framework.

\red{\subsection{MPII Cooking Activities Dataset}}
\label{sec:mpiidataset}

\red{  
The MPII Cooking Activities dataset, introduced in \citep{Rohrbach12},  contains videos of different activities in the real-world cooking domain.
Although this dataset has long and parallel action demonstrations, it provides RGB only image streams, without any depth information. Therefore, this section  investigates the performance of our SEC framework, particularly, in the case of having missing depth cues.}

\red{
We selected $5$ random cooking scenarios from this dataset and used the object tracking data provided in \cite{Yang16}. Tracking of objects was based on color and texture features processed by a random forest classifier at each 10th frame. A tracking-by-detection method \citep{Danelljan14} was further employed to complete the tracking process for missing frames. Fig.~\ref{fig:mpIIDecompImage} depicts sample frames with tracking results for one of the cooking scenarios.
Given the tracked objects, we extracted the corresponding SEC representation by simply measuring the intersection between object bounding boxes. We also introduced an additional  ``\textit{overlapping}" relation which indicates that one bounding box is completely surrounded by another one.
}

\begin{figure}[!b]
\centering
 \includegraphics[scale=0.41]{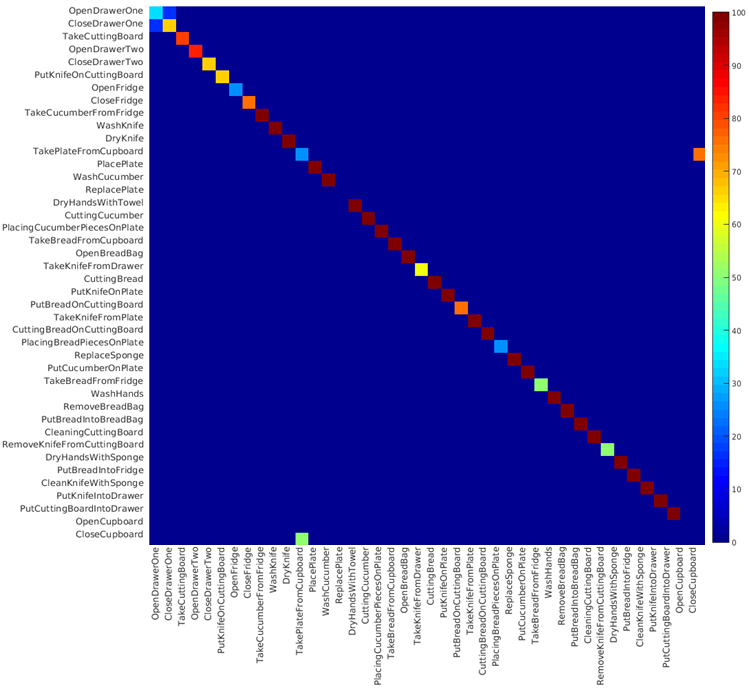}
  \caption{\red{Confusion matrix for $42$ atomic actions demonstrated in all $5$ MPII activities.}}
\label{fig:mpIIConfMatrix}
\end{figure}

\red{The selected $5$ scenarios involve in total $111$ demonstrations of $42$ different atomic actions such as  {\it Open Fridge}, {\it Cut Bread}, {\it Wash Cucumber}, {\it Dry Hands with Towel}, etc. The complete list of atomic actions is provided in Fig.~\ref{fig:mpIIConfMatrix}.
In all these $5$ scenarios, in total $20$ various objects (\eg~{\it bread, knife, fridge, etc.}) were manipulated by $4$ different subjects. There exist in total $13397$ frames in all demonstrations, \ie the average activity duration is about $1.5$ minutes. 
}

\red{
A model SEC representation for each atomic action was either extracted from their first occurrences in all demonstrations or manually introduced. These model SECs were further enriched with the object identities provided by the tracking method. 
Due to lack of depth cues, the interaction between hands and objects cannot be accurately parsed. Therefore, we applied an alternative brute force search method which scans the raw SEC matrix of each long activity and searches for the best match of model SECs by also incorporating the object information. The matched SECs are not only giving the action recognition result, but lead  also to the final temporal segmentation.  
}

\begin{figure}[!t]
\centering
 \includegraphics[scale=0.5]{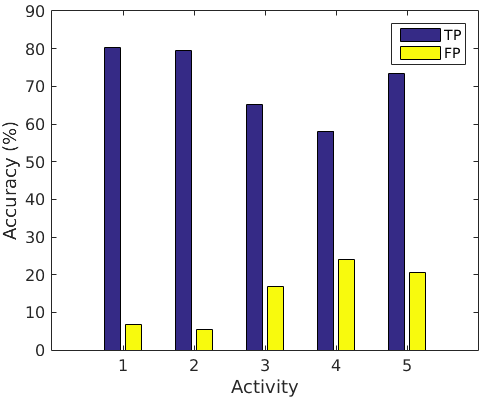}
  \caption{\red{Temporal segmentation accuracy as frame-wise true and false positive recognition rates in all $5$ MPII cooking activities.}}
\label{fig:mpIIDecompAccuracy}
\end{figure}

\red{
Fig.~\ref{fig:mpIIDecompImage} shows the frame-wise temporal segmentation result together with the human defined ground truth for one of the processed MPII cooking activity.  Each atomic action is indicated by a colored block, some of which were demonstrated in parallel. 
On the top of Fig.~\ref{fig:mpIIDecompImage} some sample frames with tracked object bounding boxes are given. 
This figure shows that our SEC framework can handle such complicated real-world sequential and parallel action demonstrations even though  depth  is not provided. The average frame-wise temporal segmentation accuracy for this activity is computed as $80\%$. 
Fig.~\ref{fig:mpIIDecompAccuracy} depicts the overall temporal segmentation accuracies as frame-wise true and false positive recognition rates in  $5$ cooking activities. The average temporal segmentation accuracy is computed as $71\%$. 
}

\red{
We also measured the class-wise recognition accuracies for each of $42$ atomic actions which were demonstrated in total $111$ times in $5$ scenarios. Fig.~\ref{fig:mpIIConfMatrix} depicts the final confusion matrix between different action types. 
The average true and false positive rates were measured as $78\%$ and $4\%$, respectively. 
In this accuracy computation step, a detected action segment is counted as true positive if there exists more than $50\%$ match with the corresponding ground truth data. Note that the false positive rate is relatively low, which also indicates that some of actions are just missed due to misaligned temporal segmentation. 
}

\begin{figure}[!t]
\centering
 \includegraphics[scale=0.45]{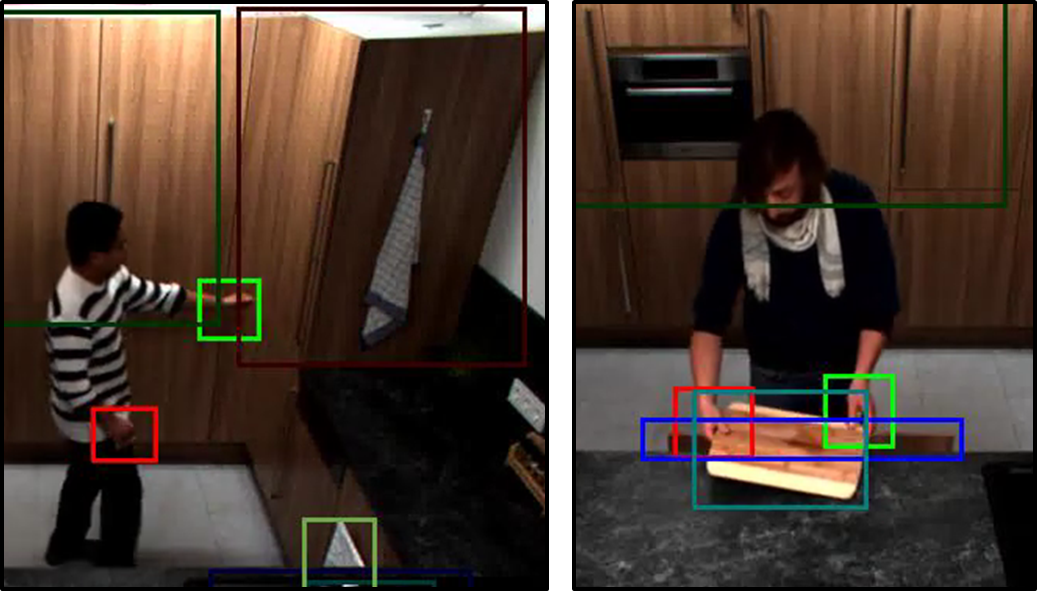}
  \caption{\red{Noisy spatial object relations due to lack of depth cues. }}
\label{fig:mpIINoisyImages}
\end{figure}

\red{
The main reason of observing slight drops  in the temporal segmentation and recognition accuracies is the missing depth information. This problem mainly  causes incorrect spatial relation computations between tracked objects. Fig.~\ref{fig:mpIINoisyImages} shows two examples from such noisy frames. On the left, the green box (\ie left hand) is touching to both dark green (\ie cupboard) and dark red (\ie fridge) boxes due to missing depth information. Note that, here, the person hand is indeed far away from both objects. On the right, bounding boxes of both hands and cutting board are still touching to the bounding box of the drawer (dark blue) even though the hand is much more above the drawer.
We, here, strongly underline the fact our SEC framework highly benefits from the depth feature of the scene. The acquired results on the MPII dataset consequently suggest that the proposed SEC framework can still provide fairly good results on real-world RGB only image streams.  
\\}

\section{Discussion}
\label{sec:discussion}

The main contribution of our paper is a novel method for automatic semantic segmentation and recognition of long and complex manipulation sequences. The proposed framework is essentially based only on the interactions between hands and manipulated objects in the scene. Our approach can consequently parse not only sequential and concurrent (overlapping) manipulation streams but also basic manipulation primitives (\eg columns of event chains) of each detected  manipulation. Without requiring any prior knowledge about objects, our framework can further extract object-like scene entities (image segments) that share similar roles in the monitored manipulations.
Furthermore, due to the fact that SECs do not care about the \emph{actually used objects}, the framework can be transferred -- without re-learning -- to different databases.

\subsection{The Problem of Action Semantics}

In Fig.~\ref{fig:action_taxonomy} we had introduced one possible taxonomy and we had started to discuss the problem of how to understand actions. Essentially, here we would like to make a case for a layered understanding: There is no such thing as a \emph{one and only action description}!  The technical literature often bypasses this problem and this may lead to biased and possibly impractical solutions. Philosophy may actually offer some useful thoughts here. For example consider a man who \emph{simultaneously} moves his arm, operates the pump, replenishes the water supply, poisons the inhabitants.  Anscombe asks whether this is one or four actions \citep{Anscombe1963}? Mele discussed this by stating (\cite{Mele1992} pg. 5):

 \begin{quote}

``Competing models of act-individuation are aligned with competing models of the individuation of events generally; and different theories of -- or at least different vocabularies for expressing -- the relationship of the mental [semantic action understanding] to the physical [sensory perception]"  (\it{square brackets added by us}).

 \end{quote}
The same problem surfaces in this study: The SEC framework cannot distinguish between ``punch" and ``push". The dynamics of these actions is very different but their SECs are the same. And the same had been observed for ``cut" and ``chop".

Mele's statement clearly expresses that such phenomena will necessarily happen for any action modeling approach. There are competing models possible for action understanding and they may capture actions at different levels of granularity!

Thus, we are allowed to understand ``cut" and ``chop" as semantically similar (destruction actions \cite{Woergoetter2013,Yang13}) but we may choose to look deeper and distinguish them -- maybe in a second stage analysis -- by ways of their different motion trajectories \citep{Woergoetter2013}.

Our own action learning supports this stance. We are first trying to master the basic skill (racket hits the ball) before learning finer motion details (topspin).

\red{The issue of hand-object relations (often “grasping”), to which much attention has been devoted \citep{Elliott1984,Cutkosky1989,Ekvall2005,Feix2009,Wimmer2011,Bullock2013,Liu2016}, is related to this semantic-level-problem, too. Also here there is often the confound that the literature mixes levels. Understanding the essence of a manipulation does, we would argue, hardly ever require an understanding of ``how it has been achieved". For example, unscrewing a lid can be performed by humans, monkeys, bears, octopuses, and some other animals all of whom employ different grasps to do so. This notwithstanding the essence of such a manipulation remains semantically the same and more such examples exist (say: the bending of a hook to make a tool done by our hands or by a Caledonian Crow, that uses its beak). This argument gets even stronger when thinking of robotic manipulators.  On the other hand, we do not wish to deny that sometimes levels blend into each other and semantically motivated (goal-directed) manipulation-requirements (for example: I want to break off a piece of rather hard wood) may indeed enforce a certain grasp (here power grasp and not pinch grip).
}

A second intriguing observation is that the SEC framework fundamentally ignores objects. Actions are extracted independently of the actual objects involved and the same physical thing (``cup") can take different roles (``being filled", ``being put on a plate", etc.). This way, the framework creates a tight link between actions and object-roles (but not objects as such) as suggested by the concept of ``Object-Action Complexes" \citep{Woergoetter2009,KrugerOAC2011}.

\subsection{Successes and Failures}

We applied our framework on three different recently published manipulation action datasets to evaluate its robustness. In each dataset, the temporal partitioning and recognition phases are quantified with respect to the human-defined ground truth. Observed high accuracies confirm the robustness of our method. One of the most fundamental advantages of our SEC based monitoring approach is the lack of the requirement of any additional training set in the case of  evaluating  novel manipulation datasets. Since SECs encode the underlying structure of manipulation actions, the already learned SEC models from our own dataset could also be employed to evaluate other datasets. This shows the generalization power of our method, which is not the case for almost all other approaches that are based on  motion patterns. We also need to emphasize that our benchmarking results are not comparable with results in \cite{Yang13} and \cite{Koo14}.  This is due to the fact that none of these benchmark providers are aimed at both, temporal segmentation and recognition of serial or parallel manipulation streams. For example, the method in \cite{Yang13} can only recognize abstract action consequences such as {\it Assemble} or {\it Transfer}. In this case, different manipulations like {\it Hiding} and {\it Putting} will be interpreted as the same class {\it Assemble}, whereas both can successfully be distinguished by our approach. As a strong contribution, our method consequently provides a richer action representation than that of others approaches. Note that the conventional datasets \citep{Schuldt04,Gupta07,Koppula13} have not been considered here since they employ  entire human body configurations and movements as main features and therefore do not provide hand-object features.

The concept of semantic event chains has also  been successfully utilized and extended by others \citep{Luo2011,Vuga2013,David2014} for monitoring purposes. The work in \cite{David2014} presented active learning of goal directed manipulation sequences, each was recognized using semantic similarities between event chains. Our scene graphs were also represented with kernels in \cite{Luo2011} to further apply different machine learning approaches. Additional trajectory information was used in \cite{Vuga2013} to reduce noisy events occur in SECs. All these studies confirm the scalability of the event chains to various monitoring tasks.

We presented our framework in a batch-type computation; that is, once the entire input stream of visual data is acquired, we first estimated the manipulator from the SEC representation and then parsed each   manipulation stream respectively. However, this is not a limitation of the proposed work, since it can also run on-the-fly, \ie  over the course of performing the activity, as soon as any kind of hand recognition method (which is not in the scope of this paper!) is additionally provided.

The main drawback of the here presented framework is the {\it segment discontinuity} problem. Since we heavily rely on tracked objects, inconsistently tracked over-segmented scenes can lead to failures in the proposed method.

\subsection{Future Directions}
To address some of the remaining problems, we are currently investigating {\it feature binding} and {\it object permanence} concepts as potential solutions to reduce failures due to the {\it segment discontinuity} problem. As discussed above, we are also aware of the fact that {\it touching} is a very unitary, discrete event. This allows rigorous classifications at a certain level of action granularity but stops sort of the finer details of an action. Consequentially, the next steps in action analysis should also involve trajectory and pose information. We strongly advocate this type a of ``layered" approach, where SECs allow classifications up to a certain semantic level and where the system can then begin ``to look deeper" allowing for further separations into finer classes. First attempts along such a layered analysis have been started \citep{WoergoetterTAMD2014} and will be much in the focus of our future works.

\section*{Appendix}

\appendix
\section{Manipulator Estimation}
\label{sec:manipulatorestimationappendix}

 \begin{algorithm}[!t]
 \caption{Computing the probability value $p_{k}$ of each object  $s_{k}$ in $\xi$.}
 \label{alg_segmentprob}
 \begin{algorithmic}
 \FORALL{object $s_{k}$ }
 \STATE $p_{k}$ = 0 (Initiation!)
 \STATE $\delta_{k}$  =  [ ]  (An empty array!)

 \FOR{r=1 to $n$ (go through all rows in  $\xi$)  }
   \IF{$s_{k}$ exists in this row!}
    \STATE $t_{Start}$ = 0  (Initiate the Start time point!)
    \STATE $t_{End}~~$ = 0    (Initiate the End time point!)

     \FOR{c=1 to $m-2$ (go through the columns!)}
      \IF{$\xi(r,c:c+1)$= $[N, T]$}
      \STATE $t_{Start}$  = c (Starting time point!)

        \FOR{f=c+2 to $m$  }
        \IF{$\xi(r,f)$= $[N]$}
        \STATE $t_{End}$  = f  (Ending time point!)
        \STATE $break$
       \ENDIF
       \ENDFOR

       \IF{$t_{End}$ $>$ $t_{Start}$}
       \STATE $\delta_{k}$ ($t_{Start}$,$t_{End}$)  =  1  (Fill the array!)
       \STATE $t_{Start}$ , $t_{End}$ = 0  (Reset time points!)
       \ENDIF

     \ENDIF
     \ENDFOR

  \ENDIF
 \ENDFOR

 \STATE $p_{k}$ = sum($\delta_{k}$)/m (Compute the final probability!)
 \ENDFOR
 \end{algorithmic}
 \end{algorithm}

In Algorithm~\ref{alg_segmentprob}, we provide the pseudocode which describes details of the {\it manipulator} estimation process in single-hand manipulation actions. This algorithm essentially describes how to compute the probability value $p_{k}$ of each object $s_{k}$ existing in the event chain  $\xi$ in order to define the likelihood of being the {\it manipulator}. The $n$ and $m$ values in Algorithm~\ref{alg_segmentprob} stand for the row and column numbers in  $\xi$.  The algorithm first searches for the start and end time points of the $[N, T, \cdots, T, N]$  sequences in all rows that include the respective object $s_{k}$ and then computes the normalized length of the touching relation $T$ to assign as the probability value. The {\it manipulator} is finally estimated as the object with the highest probability value.

\section*{Acknowledgements}
The research leading to these results has received funding from the European Community’s Seventh Framework Programme FP7/2007-2013 (Programme and Theme: ICT-2011.2.1, Cognitive Systems and Robotics) under grant agreement no. 600578, ACAT.
We thank Seongyong Koo for sharing with us the MOT dataset \citep{Koo14}.

\section*{References}


\bibliographystyle{model2-names}      

\end{document}